\documentclass[journal]{IEEEtran}
\usepackage{algorithm,algorithmic}
\usepackage{graphicx}
\usepackage{microtype}

\usepackage{booktabs} 
\usepackage{mathtools}
\usepackage{adjustbox}
\usepackage{amssymb}
\usepackage{xcolor}
\usepackage{enumerate}
\usepackage[numbers,sort&compress]{natbib}

\newcommand{\given}{\;\middle|\;}

\ifCLASSOPTIONcompsoc
    \usepackage[caption=false, font=normalsize, labelfont=sf, textfont=sf]{subfig}
\else
\usepackage[caption=false, font=footnotesize]{subfig}
\fi

\ifCLASSINFOpdf
\else
\fi
\usepackage{algorithm}
\usepackage{algorithmic}
\usepackage{amsmath}
\usepackage{graphics}
\usepackage{epsfig}

\begin{document}
%
\title{graph}
%
%
%

\author{Jiawei Wang and  Lijun Sun*
\thanks{The authors are with the Department of Civil Engineering, McGill University, Montreal, QC H3A 0C3, Canada. \textit{(Corresponding author: Lijun Sun.)}}
}%

\newcommand{\citeay}[1]{\citeauthor{#1}\,\shortcite{#1}}

\title{Robust Dynamic Bus Control: A Distributional Multi-agent Reinforcement Learning Approach}


\maketitle

\begin{abstract}
Bus system is a critical component of sustainable urban transportation. However, the operation of a bus fleet is unstable in nature, and bus bunching has become a common phenomenon that undermines the efficiency and reliability of bus systems. Recently research has demonstrated the promising application of multi-agent reinforcement learning (MARL) to achieve efficient vehicle holding control to avoid bus bunching. However, existing studies essentially overlook the robustness issue resulting from various events, perturbations and anomalies in a transit system, which is of utmost importance when transferring the models for real-world deployment/application. In this study, we integrate implicit quantile network and meta-learning to develop a distributional MARL framework---IQNC-M---to learn continuous control. The proposed IQNC-M framework achieves efficient and reliable control decisions through better handling various uncertainties/events in real-time transit operations. Specifically, we introduce an interpretable meta-learning module to incorporate global information into the distributional MARL framework, which is an effective solution to circumvent the credit assignment issue in the transit system. In addition, we design a specific learning procedure to train each agent within the framework to pursue a robust control policy. We develop simulation environments based on real-world bus services and passenger demand data and evaluate the proposed framework against both traditional holding control models and state-of-the-art MARL models. Our results show that the proposed IQNC-M framework can effectively handle the various extreme events, such as traffic state perturbations, service interruptions, and demand surges, thus improving both efficiency and reliability of the system.
\end{abstract}

\begin{IEEEkeywords}
Bus bunching, robust holding control, multi-agent reinforcement learning, distributional reinforcement learning, meta-learning.
\end{IEEEkeywords}

%
\IEEEpeerreviewmaketitle

\section{Introduction}
%
%
%
%

\IEEEPARstart{T}{he} operation of bus systems is unstable in nature due to the great uncertainties in traffic states and passenger demand. Bus bunching has been a long-standing operation problem resulting from these uncertainties. If a bus is slightly delayed on the route, it will be more likely to encounter more waiting passengers at the next bus stop, and then the dwell time serving boarding/alighting passengers will also increase. As a result, a delayed bus will be further delayed on a route. Due to the instability of the system, bus bunching has become the most critical operational issue that undermines the reliability and efficiency of bus services.
With recent advances in artificial intelligence \citep{mnih2015human,silver2017mastering}, researchers have identified reinforcement learning (RL) as a promising solution to solve complex traffic control problems, such as traffic signal control \citep{li2016traffic,chu2019multi} and ramp metering control \citep{belletti2017expert}. To address the bus bunching problem, recent studies have also modeled bus holding control of a bus fleet in RL and multi-agent reinforcement learning (MARL) framework \citep[see e.g.,][]{chen2016real,alesiani2018reinforcement,wang2020dynamic}. However, modeling bus holding control in MARL is a challenging problem. First, there exist some common and tricky challenges in learning robust control policies in MARL, such as the non-stationary dynamics and the complex credit assignment issue. Second, bus holding is an asynchronous control problem---a bus can only implement the holding action when it arrives at a bus stop. As a result, we cannot directly adapt existing MARL frameworks to model bus fleet operation. To address the asynchronous issue, \citet{wang2021reducing} propose an inductive critic to improve credit assignment among agents in an asynchronous setting and the model has demonstrated superior performance in learning bus holding control policies to avoid bus bunching. Despite the promising results, there still exist several critical challenges in adapting the RL for real-world operations/deployments.
On the one hand, the credit assignment may not require an explicit formulation \citep{LearningImplicitCreditAssignment20}, since approximating the contribution of each agent may introduce an approximation error that hinders policy training. {On the other hand, previous RL-based fleet control methods have not fully examined the robustness of the solutions, which is of utmost importance for real-world applications. As a result, these control policies may behave less efficiently under unexpected scenarios, such as demand impulse at some stops (i.e., there is a special event in the neighborhood) and incidents causing traffic congestion.} To fill these gaps, in this study we propose a distributional MARL framework to learn robust bus holding control policies. In particular, we introduce an interpretable meta-learning module to address the aforementioned non-stationary dynamics and complex credit assignment issues when modeling bus fleet as a multi-agent system. Our main contributions can be summarized as follows:
\begin{itemize}
  \item To the best of our knowledge, this study is the first to introduce distributional MARL to handle uncertainties and achieve robustness in bus fleet operations and holding controls.
  \item We design a meta-learning-based framework to adapt distributional RL for efficient multi-agent control policy learning.
  \item We design a training procedure for learning robust holding control policies in the distributional MARL framework.
  \item We evaluate the performance---in terms of both effectiveness and robustness---of the proposed framework through extensive simulation experiments on four real-world bus routes.
\end{itemize}

The rest of the paper is organized as follows. Section~\ref{sec:related_work} presents the bus fleet holding control problem and discusses related works on traditional optimization-based models and recent RL- and MARL-based solutions. Section~\ref{sec:background} introduces the background and preliminary knowledge of this study. Section~\ref{sec:method} presents the proposed distributional MARL approach to model bus fleet control. Experimental studies are presented in Section~\ref{sec:experiments} to evaluate the proposed model. Finally, Section~\ref{sec:conclusion} summarizes some  concluding remarks and discusses future research directions.





\section{Related Work} \label{sec:related_work}

\subsection{Research on Bus Control Strategy}
Bus holding control has long been considered the most effective solution to reduce bus bunching, because it can make a good compromise among control efficiency, implementation practicability, and service satisfaction of transit users \cite{eberlein2001holding,cats2011impacts,wu2017modelling,wang2020dynamic}. According to the previous studies, there are mainly two traditional approaches to develop holding control strategies with respect to the solution method. The first is the optimization-based approach \cite{hickman2001analytic,eberlein2001holding,delgado2009real}, which mainly focuses on deriving holding period/duration on a rolling horizon analytically by optimizing objectives related to system efficiency such as passenger waiting time. These studies need to model the future states of the system and implement optimization at each decision stage, which turns out to be less efficient. The second approach is to develop rule-based holding control strategies based on the linear control law \cite{xuan2011dynamic,berrebi2015real,daganzo2009headway}. These studies are in general computationally efficient but may be less reliable due to the simple linear assumptions in reproducing the complex dynamics in transit operations. Notably, most recent studies have identified that incorporating real-time information is a key solution to efficient transit control. For example, \citet{berrebi2018comparing} evaluated the holding control methods and concluded that predictive control based on real-time information could achieve the best compromise between headway regularity and holding cost. \citet{gkiotsalitis2020analytic} introduced an analytic solution for the single variable bus holding problem that considers the real-time passenger demand and vehicle capacity limits. \citet{wang2020dynamic} designed MARL bus fleet control framework considering real-time global control decisions. These studies demonstrate that real-time information can offer more efficient and more effective control policies.

Uncertainties in transit demand (i.e., passenger arrival rate) and traffic state (i.e., on-road running time) are two most important factors that undermine the robustness and reliability of transit operation. Recent studies have taken these uncertainties into account in building analytical models. For example, \citet{xuan2011dynamic} considered stochastic on-road running time in developing robust control holding strategies.
\citet{wu2017modelling} proposed adaptive control strategies that take into account dynamic redistribution of passenger queues.
\citet{van2019robust} developed an assessment framework to systematically assess the quality of a transit service under headway-based holding strategies. Notably, they evaluate the control performance with two designed scenarios: trip detour and traffic disruption.
\citet{li2019robust} derived robust control strategies for schedule adherence and headway regularity considering both the randomness in demand and traffic state.
Although these studies have contributed to developing robust bus control, there are still some major limitations. On the one hand, most previous studies only introduce perturbations in demand and speed as a way to incorporate uncertainties; however, these scenarios are essentially easy to solve and should be categorized as normal scenarios in daily transit operations. A more challenging task is actually to adapt for extreme scenarios and anomalous events, such as service disruptions, on-road incidents, and demand surges, which may lead to cascading failures in the system. On the other hand, existing models cannot effectively learn from the past to update the control policy, as they often require an exact mathematical model of the complex transit system with a prior setting. Therefore, they become less efficient in handling the unstable dynamic of the transit system.

\subsection{Research on Transit Control with Reinforcement Learning}
RL provides an end-to-end control learning paradigm over traditional methods, allowing model-free control and continuous learning with long-term consideration \cite{sutton2018reinforcement}. Thanks to the growing computation ability, researchers have extended RL to DRL by exploiting deep neural networks \cite{mnih2015human, lillicrap2015continuous}. Furthermore, MARL, which is the DRL scheme for the multi-agent case, provides an efficient framework to explore a cooperative control solution.
Recent years have witnessed increasing research on DRL / MDRL contributing to the transit control problem. For example, \citet{alesiani2018reinforcement} implemented holistic holding control based on DRL. \citet{chen2016real} and \cite{wang2020dynamic} utilized MARL framework to achieve decision coordination among buses in the transit system. \citet{wang2021reducing} proposed an MDRL framework to achieve efficient holding control by addressing the asynchronous issues in transit operation. All these studies have demonstrated the superior performance of DRL/MDRL on vehicle holding control. However, before having real-world deployments, it remains an open question how to achieve robust control under uncertainty and extreme scenarios with RL based control scheme.

\section{Background} \label{sec:background}
\subsection{Reinforcement Learning based Control}
RL provides an adaptive scheme to explore optimal control policy in a system that is governed by a Markov decision process (MDP). Specifically, the learner and controller in this system is referred to as the agent, while all the other components are regarded as the environment \cite{sutton2018reinforcement}.
At each decision step $t$, the agent decides an action $a_t \in \mathcal{A}$ based on its current policy $\pi_{t}$, given a state $s_t \in \mathcal{S}$ observed from the environment. Then, the agent moves to the next state $s_{t+1} \in \mathcal{S}$, and receives reward signal $r_t \in \mathcal{R}$ as a result of this interaction. Based on the MDP assumption, the dynamic of this process can be represented by the transition probability: $p\left(s',r \given s ,a \right) \doteq \mathrm{Pr}\left\{s_{t+1}=s',r_t=r \given s_t = s, a_t = a\right\}$.

In the RL task, the goal of the agent is to learn an optimal policy $\pi^*$ to maximize the cumulative reward signal along the control horizon $H$. Generally, the cumulative reward signal can be formulated as the expected return with a discounted factor $\gamma \in \left[0,1\right]$:
\begin{equation}
R_t \doteq \sum^H_{k=0} \gamma^k r_{t+k}.
\label{eq:r}
\end{equation}

There are mainly two categories of RL methods---value-based methods and policy-based methods. Value-based methods, such as Q-learning and Deep Q-Networks (DQN) \cite{mnih2015human}, use state-action value function $Q\left(s,a\right)$ for policy evaluation. The basic idea is that after achieving a good estimate of the optimal state-action value $Q^*\left(s,a\right)$, the optimal policy $\pi^*$ will be the one choosing action $a$ with the highest state-action value at state $s$. Most RL methods estimate the state-action value function by using the Bellman equation as an iterative update \cite{bellman1957markovian}. This scheme is also referred to as bootstrapping \cite{sutton2018reinforcement}, i.e., updating the state-action value estimate for a state based on the estimated values $\overline{Q}$ of subsequent states:
\begin{equation}
Q\left(s_{t},a_{t}\right) = \mathrm{E}_{s_{t+1}}\left[r_t + \gamma \max_{a_{t+1}}\overline{Q}\left(s_{t+1},a_{t+1}\right)\right].
\label{eq:Q}
\end{equation}
As an extension of Q-learning, DQN introduces deep neural network to learn/approximate $Q^*$ by minimizing the following loss function:
\begin{equation}
\mathcal{L}\left(\theta\right) = \mathrm{E}_{s,a,r,s'}\left[Q^*(s,a)-\left(r+\gamma \max_{a'}\overline{Q}\left(s',a'\right)\right)\right].
\label{eq:DQN}
\end{equation}


Policy-based methods, such as policy gradient (PG) \cite{sutton1999policy} and deterministic policy gradient (DPG) \cite{silver2014deterministic}, parameterize policy as $\pi_{\theta}$ and then directly optimize $\pi_{\theta}$ to achieve a larger cumulative reward. Formally, the objective of these methods is to maximize the cumulative reward  $\mathcal{J}\left(\theta\right)=E_{s_t,a_t}\left[R_t\right]$. In PG, the action is sampled from a parametric probability distribution $a_t \sim \pi_{\theta}\left(a_t \given s_t \right) = P\left(a_t \given s_t,\theta \right)$. The policy parameters are updated through gradient ascent with respect to the cumulative reward:
\begin{equation}
\nabla_{\theta}J(\theta)=\mathrm{E}_{s_t,a_t \sim \pi_{\theta}\left(\cdot \given s_t\right) }\left[\nabla_{\theta}\log \pi_{\theta}\left(a_t\given s_t\right) G^{\pi}\left(s_t,a_t\right)\right].
\label{eq:pg}
\end{equation}
As PG adopts a stochastic policy, it suffers from large variances during the learning process. To address this issue, DPG suggests a deterministic mapping from state to action: $a_t = \pi_{\theta}\left(s_t\right)$. By doing so, we only need to sample from the state space to update the parameters, and the gradients can be calculated as:
\begin{equation}
\nabla_{\theta}J\left(\theta\right)=\mathrm{E}_{s_t}\left[\nabla_{\theta}\pi_{\theta}\left(s_t\right) \nabla_{a_t}G^{\pi}\left(s_t,a_t\right)|_{a_t = \pi_{\theta}\left(s_t\right)}\right].
\label{eq:dpg}
\end{equation}

In general, the cumulative reward $R_t$ can be estimated by learning the state-action value function $Q\left(s, a\right)$. Such a framework is referred to as actor-critic \cite{sutton2018reinforcement}, in which the state-action value estimator is called the critic and the policy is modeled on the actor.

\subsection{Distributional Reinforcement Learning.}
Different from traditional RL methods that learn the expectation of the cumulative reward or the state-action value, distributional RL explores the state-action value distribution and thus allows us to account for the intrinsic uncertainty of an MDP. In distributional RL, the state-action value distribution can be modeled as a distributional Bellman equation \cite{bellemare2017distributional}:
\begin{equation}
Z\left(s_t,a_t\right) \doteq r\left(s_t,a_t\right)+\gamma Z\left(s_{t+1},a_{t+1}\right),
\label{eq:d_bellman}
\end{equation}
where the state-action value distribution $Z\left(s_t,a_t\right)$ takes into account the uncertainty from reward signal $r\left(s_t,a_t\right)$, the transition to the next state $\left(s_{t+1},a_{t+1}\right)$, as well as the next random return $Z\left(s_{t+1},a_{t+1}\right)$.

To facilitate state-action value distribution learning within traditional DRL frameworks (e.g., DQN), a state-of-the-art solution is to approximate the target distribution with quantile function and approximate the state-action value \citep{dabney2018implicit} by:
\begin{equation}
\begin{split}
Q\left(s_t,a_t\right)&=\mathrm{E}_{\tau\sim \mathcal{U}\left[0,1\right]}\left[Z_{\tau}\left(s_t,a_t\right)\right] \\
&\approx \sum^{N-1}_{i=0} \left(\tau_{i+1}-\tau_{i}\right)Z_{\hat{\tau}_i}\left(s_t,a_t\right),
\end{split}
\label{eq:iqn}
\end{equation}
where $\hat{\tau}=\frac{\tau_{i+1}+\tau_{i}}{2}$ and $Z_{\tau}\left(s_t,a_t\right)$ represents the quantile function at fraction $\tau \in \left[0,1\right]$ for $Z\left(s_t,a_t\right)$. Then, we can sample $\tau$ from a uniform distribution $\mathcal{U}\left[0,1\right]$ to produce a state-action value sample of $Z\left(s_t,a_t\right)$.

According to previous studies \citep{bellemare2017distributional,dabney2018implicit}, distributional RL exhibits superior performance in terms of both training convergence and execution performance. By modelling the value distribution, we can better characterize the intrinsic uncertainties in MDP and thus identify more robust and safer control policies \citep{2021riskaverse}.

\subsection{Distorted State-action Value Distribution}

The distorted expectation can be expressed as a weighted average of quantiles \citep{dhaene2012remarks}. Specifically, given a distortion function $g:\left[0,1\right]\rightarrow \left[0,1\right]$ satisfying $g\left(0\right)=0$ and $g\left(1\right)=1$, the distorted expectation of state-action value distribution $Z$ is $\int_{0}^{1} Z_{\tau}\mathrm{d} g\left(\tau\right)=\int_{0}^{1}g'\left(\tau\right) Z_{\tau}\mathrm{d} \tau $, where $g'\left(\tau\right)$ is the derivative of $g$ and it can be considered the distortion weights for quantiles. For example, \citet{wang2000class} proposed the following distortion function:
\begin{equation}
 g\left(\tau\right)=\Phi\left(\Phi^{-1}\left(\tau\right)+\beta\right) ,
 \label{eq:distort}
\end{equation}
where $\Phi$ represents the cumulative distribution function (CDF) of standard normal distribution, and $\Phi^{-1}$ is the inverse CDF; and $\beta>0$ corresponds to being risk-averse while $\beta<0$ indicates being risk-seeking. Generally, the distorted state-action value expectation can be approximated by:
\begin{equation}
\hat{Q}\left(s_t,a_t\right)=\sum^{N-1}_{i=0} \left(\tau_{i+1}-\tau_{i}\right)g'\left(\hat{\tau}_i\right)Z_{\hat{\tau}_i}\left(s_t,a_t\right).
\label{eq:diqn}
\end{equation}

\begin{figure}[!ht]
\centering
  \includegraphics[scale=0.25]{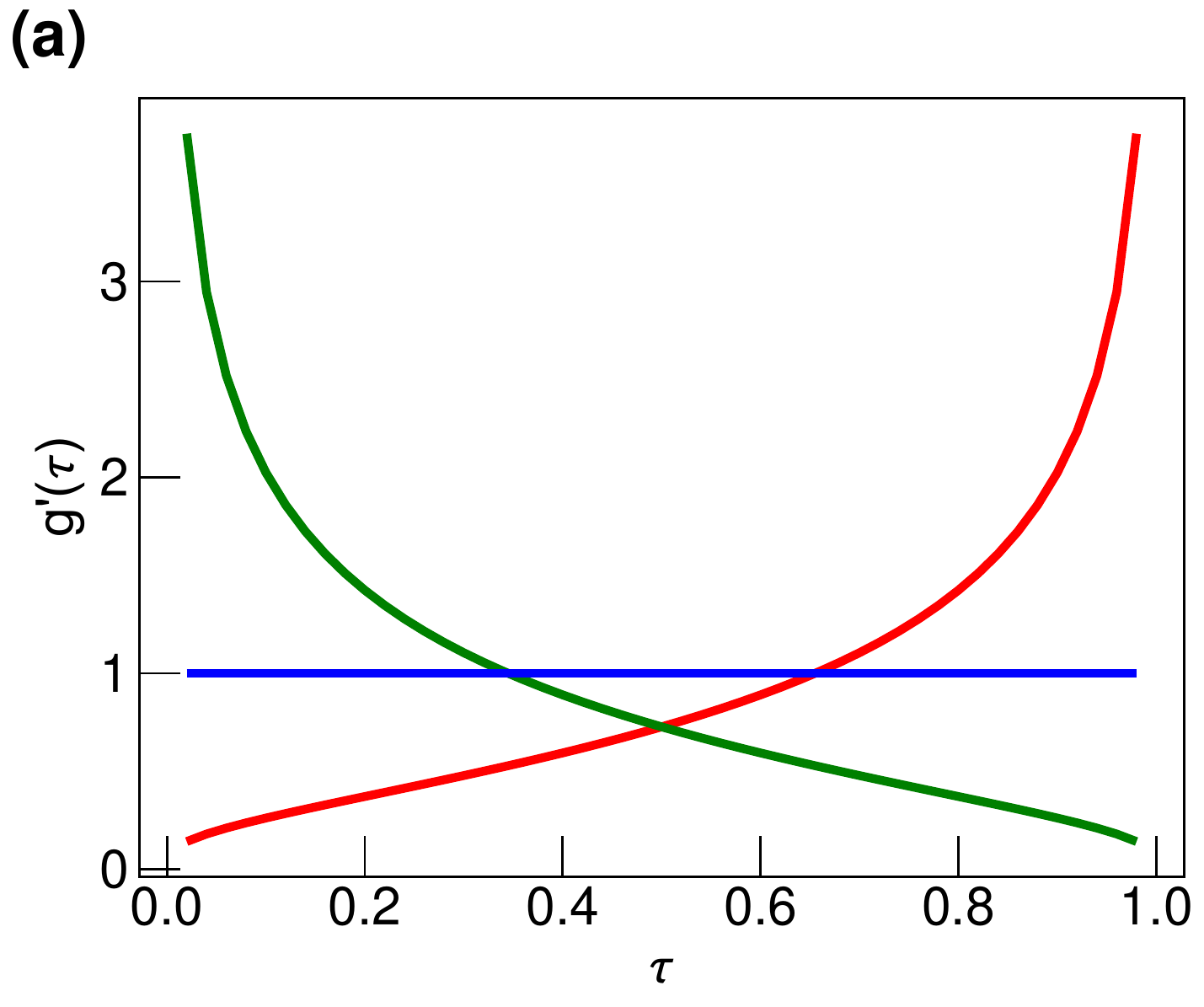} \qquad
\includegraphics[scale=0.25]{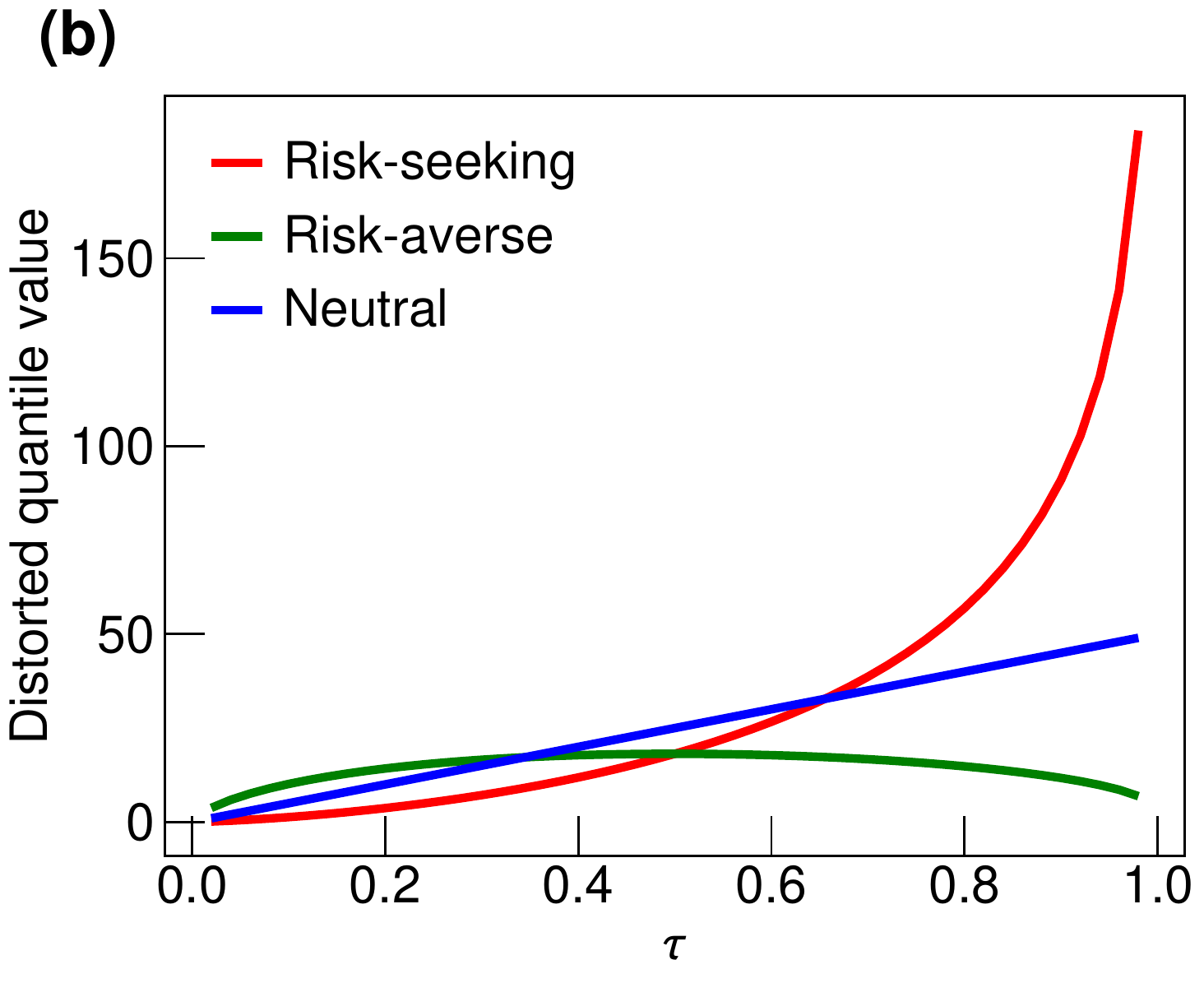}
 \caption{(a) Distortion weights from $g'\left(\tau\right)$  where we set $\beta=0.8$ for risk-averse preference and $\beta=-0.8$ for risk-seeking preference; (b) Distorted state-action value illustration based on $g'\left(\tau\right)$.}
  \label{fig:dd}
\end{figure}

As shown in Fig.~\ref{fig:dd}, the distortion weight essentially corresponds to the preference on policies by distorting state-action value distribution. For example, the risk-averse policy generally places lower weight on larger quantiles.

\subsection{Meta Reinforcement Learning}

Meta RL introduces meta-learning to improve learning performance \cite{metazhong2018} and make it easier for agents to generalize to new tasks \cite{Abhishek2018generalize}. Specifically, meta-gradient RL \cite{metazhong2018} is one of the most popular schemes for optimizing hyper-parameters in RL, which tunes the differentiable ingredients in the RL pipeline using a higher-order gradient. In general, the RL parameters $\theta$ (i.e., parameters of the state-action value approximator or the policy) are updated as follows:
\begin{equation}
\begin{split}
\theta' &= \theta + f\left(d,\theta,\eta\right), \\
f &=\frac{\partial \mathcal{J}}{\partial \theta},
\end{split}
\label{eq:meta-expression}
\end{equation}
where $J$ is the objective function (e.g., Eq.~\eqref{eq:DQN} ) for the RL agent and $\eta$ represents the hyper-parameter for meta learning. We sample trajectories $d$ (e.g., $\left(s,a,r,s'\right)$) to calculate gradient for $J$. In meta-gradient RL, $\eta$ is updated based on the gradient from the derivative of a differentiable meta-objective $J'$. Specifically, the derivative is calculated with another sample trajectory $d'$ and a fixed $\eta'$ as a reference value:
\begin{equation}
\frac{\partial {\mathcal{J}'\left(d',\theta',\eta'\right)}}{\partial \eta} =\frac{\partial \mathcal{J}'\left(d',\theta',\eta'\right) }{\partial \theta'}\frac{\partial \theta'}{\partial \eta}.
\label{eq:meta-update}
\end{equation}
Intuitively, after every update on $\theta$, meta RL adapts the meta-parameters $\eta$ to the direction that achieves better performance.


\section{Methodology} \label{sec:method}
\subsection{Bus Control Framework}

\begin{figure}[!t]
\centering
     \includegraphics[scale=0.35]{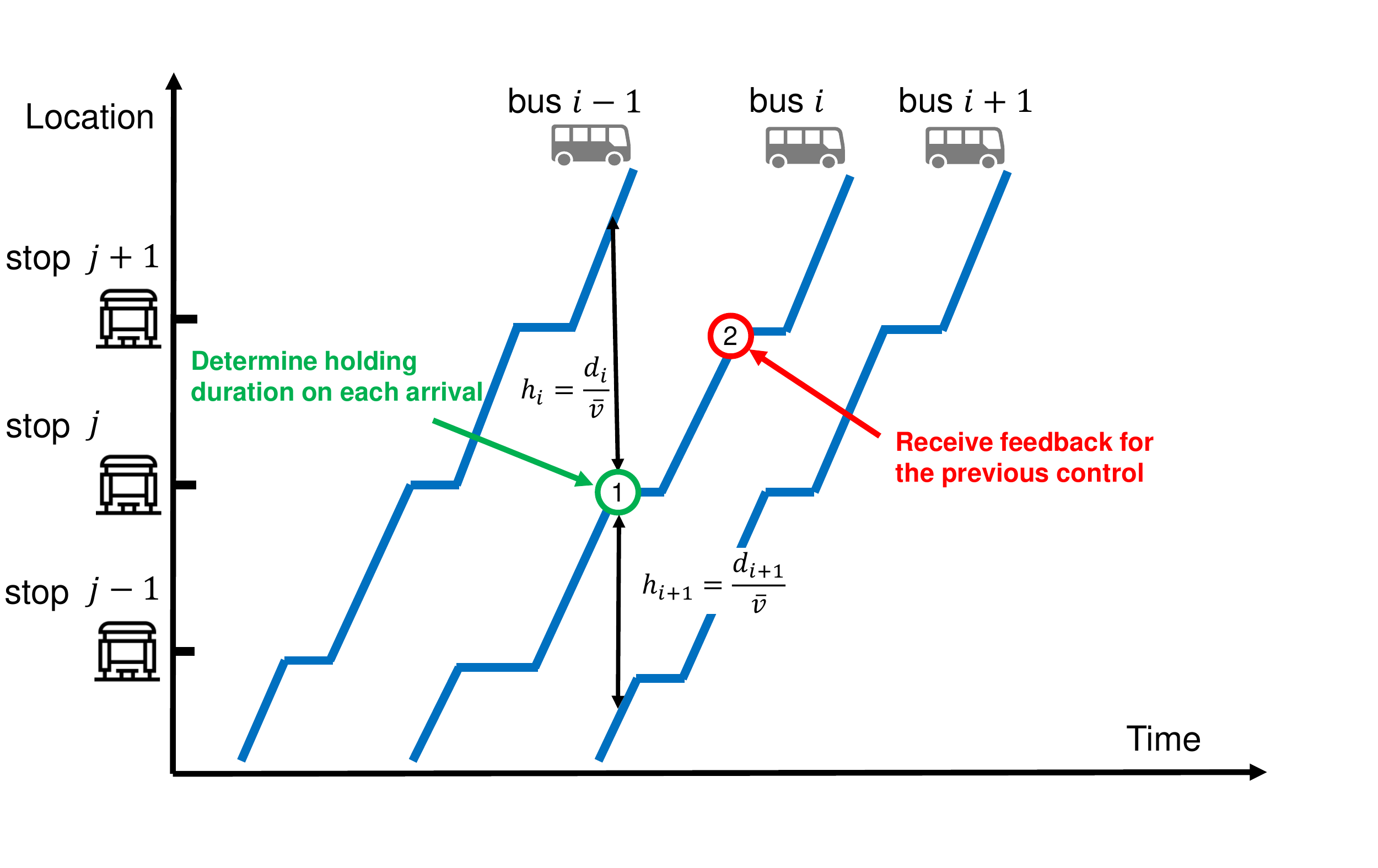}
  \caption{The vehicle control illustration on a bus route. When bus $b_i$ arrives at a bus stop, we denote by $h_{i+1}$ the backward headway (time for the following bus $b_{i+1}$ to reach the current location of $b_i$) and $h_{i}$ the forward headway (time for $b_i$ to reach the current location of $b_{i-1}$). The bus has access to the number of onboard passengers and it also observes the number of waiting passengers at the bus stop. Each bus will determine holding period/duration once it arrives at a stop (i.e., the green circle) and receive feedback when arriving at the next stop (i.e., the red circle). }
   \label{fig:framework}
\end{figure}

Bus holding control is the most widely used strategy to maintain consistent headway and avoid bus bunching in daily operation \citep{cats2011impacts}. The essential idea of holding control is to let a bus stay longer (i.e., adding a slack period) at a bus stop in addition to the dwell time for passengers to board/alight. In this study, we model bus holding as a control policy in MARL with the objective of improving operational efficiency. As shown in Fig.~\ref{fig:framework}, all buses in the transit system follow a sequential order, where vehicle $b_{i+1}$ follows vehicle $b_i$.

In \citet{wang2021reducing}, bus fleet operation and holding control are formulated as a Markov game \cite{littman1994markov}. The agent (i.e., bus) implements an action (i.e., determine holding period) once it arrives at a bus stop and receives feedback when arriving at the next bus stop. In particular, the authors model bus holding control as an asynchronous control task, in which it is not necessary for all agents to act simultaneously. Following the same framework as in \cite{wang2021reducing}, we consider a partially observable scenario in which each bus asynchronously determines the holding duration based on local observations.
For a transit system with $N$ buses, each bus maintains its local state transition: $\hat{\mathcal{P}_i}:\mathcal{S}_i\times\hat{\mathcal{A} }\mapsto\mathcal{S}_i, \mathcal{S}_i \subseteq \mathcal{S}, \hat{\mathcal{A}} \subseteq \mathcal{A}$, where $S_i$ is the observation space of agent $i$,  $\mathcal{S}=\left\{S_1,\ldots,S_N\right\}$ and $\mathcal{A}=\left\{A_1,\ldots,A_N\right\}$ denote the global state  space and the joint action space, respectively. The parametrized policy $\pi_{theta}$ chooses an action $a_i$ based on local observation $s_i$, i.e., $a_i = \pi_{\theta_i}\left(s_i\right)$. Finally, the reward function for each agent can also be defined independently: $\mathcal{R}=\left\{R_1, \ldots,R_N\right\}$. In the following, we define the elements in the RL pipeline for bus fleet operation on a single route. We refer to each ``bus'' as an ``agent'' and use the two terms interchangeably.

\subsubsection{State}

For each agent, we refer to its local observation of the system as the state. For bus $b_i$ arriving at bus stop $k_j$ at time $t$, we denote its state by $s_{i,t}$. It should be noted that the definition of state plays an important role in the overall RL framework. In order to support reliable and robust decision making, the elements of state should be sufficiently rich in characterizing the complex dynamics of the system. In the meanwhile, the state should be accessible---something that can be collected from available sensors on a bus---for real-world and real-time implementations. In this study, we consider two components in defining states for a bus $b_i$: the first component consists of the  forward headway $h_i$ and the backward headway $h_{i+1}$; the second component focuses on passenger demand, and we use the number of onboard passengers and the number of waiting passengers at the bus stop to define it. In practice, the first component on vehicle information can be measured in real-time through the automatic vehicle location (AVL) system, and the second component on passenger demand can be obtained from the automatic passenger counting (APC) system and the automated fare collection (AFC) system.



\subsubsection{Action}

Following \cite{wang2020dynamic}, we model holding time as:
\begin{equation}\label{eq:action}
\triangle d^{t}_i=a_{i,t} \triangle T,
\end{equation}
where $\triangle T$ is the maximum holding duration and $a_{i,t} \in \left[0,1\right]$ is a strength parameter. We consider $a_{i,t}$ the action of bus $b_i$ when arriving at a bus stop at time $t$. Here, $\triangle T$ is used to restrict the maximum holding duration and avoid over-intervention, and we model $a_{i,t}$ as a continuous variable to explore the near-optimal policy in an adequate action space. No holding control is implemented when $a_{i,t}=0$.

 \subsubsection{ Reward} Although holding control can reduce the variance of bus headway and therefore promote system stability, in the meanwhile, it also introduces additional slack time at stops. As a result, passenger travel time and route operation time (i.e., the total duration from leaving the departure terminal to arriving at the final destination) will also increase, imposing an additional penalty on passenger travel cost and also requiring a larger fleet to maintain service frequency. To balance system stability and operation efficiency, we design the reward function associated with bus $b_i$ at time $t$ as:
\begin{equation}
{r_i^t}= -\left(1-w\right)\times CV^2-w\times a_{i,t},
\label{eq:reward}
\end{equation}
where $CV^2=\frac{\mathrm{Var}\left[H\right]}{{\mathrm{E}}^2\left[H\right]}$ is an indicator for headway variability \cite{NAP24766}. Essentially, ${\mathrm{E}}\left[H\right]$ is almost a constant based on the schedule, and thus $CV$ is mainly determined by $\mathrm{Var}\left[H\right]$: a small $CV$ indicates consistent headway values on the bus route, and a large $CV$ suggests heavy bus bunching. The second term in Eq.~\eqref{eq:reward} penalizes holding decisions and prevents the learning algorithm from making excessive control decisions. Overall, the goal of this reward function is to achieve system stability with as few interventions as possible, with $w$ being a balancing parameter. We set $w=0.2$ in this study.

\subsection{Distorted Distributional Multi-agent Reinforcement Learning for Robust Transit Control}
\begin{figure}[!ht]
\centering
   \includegraphics[scale=0.25]{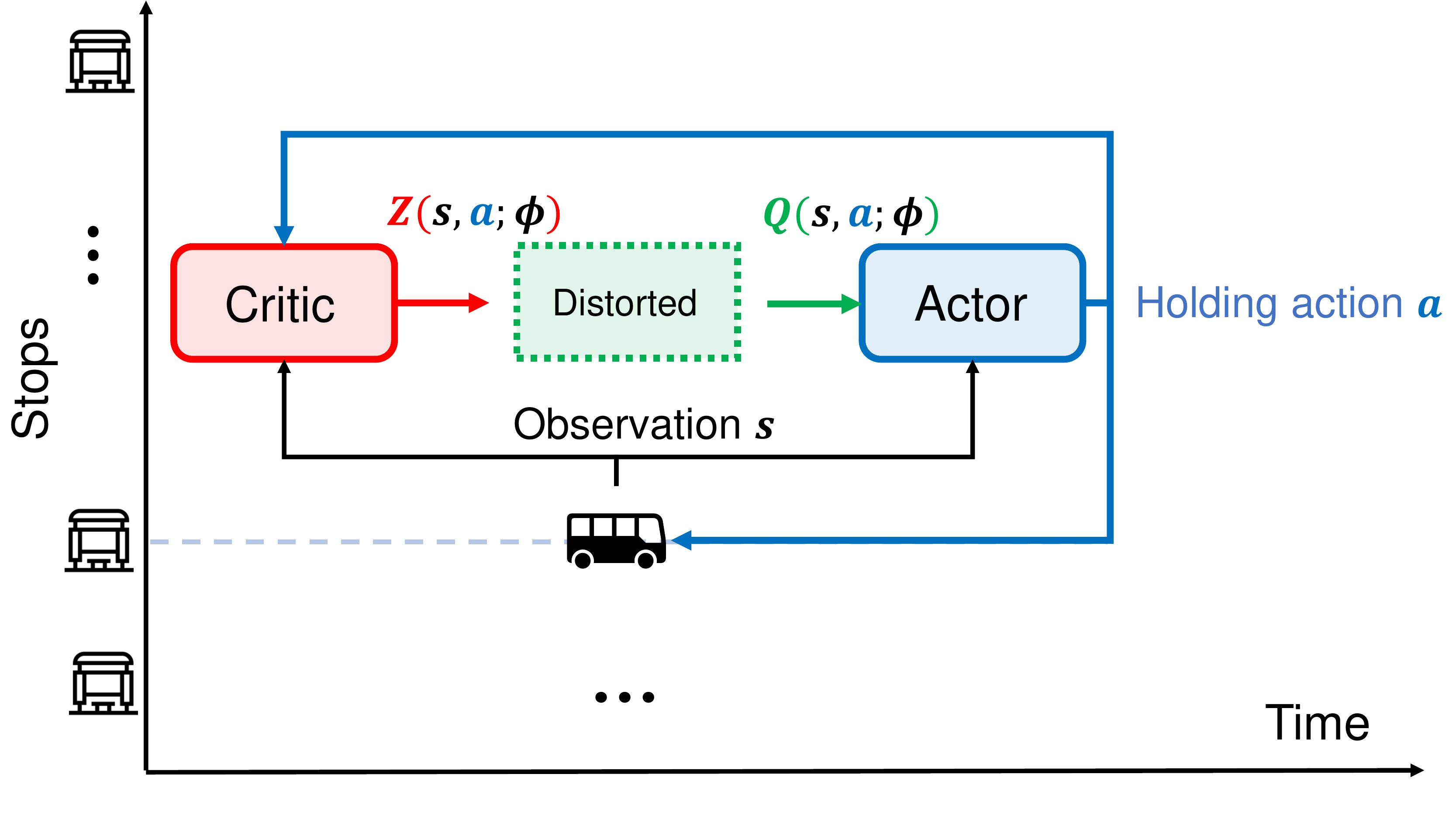}
 \caption{Illustration of training and execution of a bus in operation based on distorted distributional MARL: Actor determines action to interact with the environment (i.e., transit system). Critic evaluates the overall impact from the action based on observation. This evaluation is further distorted considering the uncertainty from the transit system operation and then used for policy optimization.}
  \label{fig:idea}
\end{figure}

Modelling uncertainty of transit operation using probability distribution is one of the most common solutions to derive robust control policies \citep{yin2014online,khadilkar2018scalable}. While distributional RL provides an efficient framework to consider uncertainty by learning state-action value distribution, there still exist two critical challenges in exploiting distributional RL in a multi-agent transit system. The first is the non-stationary environment as a result of the unstable dynamics of the transit system and the undetermined behavior of each bus; the second is the credit assignment resulting from the undetermined contribution of each agents' decisions during real-time operation.
Motivated by the risk-sensitive RL, which uses the distorted state-action value distribution to learn policies with risk preference \citep{risaverse2021}, we manage to address the above challenges by adapting this idea to distributional MARL.
In this study, we adopt actor-critic \citep{sutton2018reinforcement} as basic architecture. Specifically, each agent will maintain an actor network to model policy and a critic network for policy evaluation. As shown in Fig.~\ref{fig:idea}, we consider distributional critic that learns the full state-action value distribution and therefore takes into account the uncertainty in the transit system as well as the behavior of other agents. For the actor that determines holding control, we train it based on the distorted state-action value distribution from the critic. Particularly, we treat risk preference as confidence level in our multi-agent setting.  For example, the  risk-seeking preference of an agent can be interpreted as high confidence in the contribution of its own holding control during the operation horizon. By imposing different confidence levels, we can produce distorted state-action value distributions to reflect agents' confidence levels in their contributions. By doing so, the model can avoid approximation error during credit assignment and take into account the uncertainty during the agents' policy evaluation. In the following, we will introduce the methodology in detail.

 \subsubsection{Distributional Critic to Handle Uncertainty}
In this study, the critic learns to evaluate the control decision of each bus. To handle the uncertainty from the transit operation, we use a distributional critic, which models quantile function for state-action value for each state-action pair $(s_t,a_t)$. It should be noted that in general independent actor-critic frameworks, the critic only takes local observations for policy evaluation. In order to incorporate global information to promote efficient policy evaluation, we propose a meta-gradient learning scheme in Section~\ref{subsubsec:meta}. Specifically, state-action value at quantile fraction $\tau$ is $Z_{\tau}\left(s_t,a_t;\phi\right)$, where $\phi$ denotes the parameters of the critic.

Traditionally, the critic learns state-action value expectation by minimizing temporal-difference (TD) error \citep{sutton2018reinforcement}. \citet{dabney2018implicit} considered distributional RL and adopted quantile functions to approximate the state-action value distribution. The TD error between two state-action values, on quantile fractions $\tau_{k}$ and $\tau_{k'}$, is:
\begin{equation}
\delta_{kk'}^t = r_t+\gamma\left[ Z_{\tau_{k'}}\left(s_{t+1},a_{t+1};\phi^-\right) \right] - Z_{\tau_{k}}\left(s_{t},a_{t};\phi\right) ,
\label{eq:td}
\end{equation}
where $\phi^-$ denotes the parameter in the target network \citep{mnih2015human}. Following the implicit quantile network (IQN) \citep{dabney2018implicit}, quantile fractions are randomly sampled from a uniform distribution (i.e., $\tau_k, \tau_{k'} \sim \mathcal{U}\left(0,1\right)$) and fed into the critic after embedding.

To formulate an efficient objective to train the distributional critic, previous studies suggested using Huber loss  \citep{huber1992robust} due to smooth gradient and proposed to use quantile Huber loss as a surrogate for the Wasserstein  distance between two one-dimensional distributions \citep{dabney2018implicit}. Specifically, the Huber loss \citep{huber1992robust} with threshold $\kappa$ is defined as:
\begin{equation}
\mathcal{L}_{\kappa}\left(x\right) =
    \begin{cases}
      \frac{1}{2}x^2 & \text{if }{ \left| x \right| < \kappa }, \\
      \kappa\left(\left| x\right|-\frac{1}{2}\kappa\right) & \text{otherwise} ,
    \end{cases}
\label{eq:huberloss}
\end{equation}
 where $x$ refers to the residuals.
 The quantile Huber loss at quantile fraction $\tau$ is:
\begin{equation}
\rho^{\kappa}_{\tau}\left(x\right)=\left|\tau- \mathbb{I}\left\{x<0\right\}\right|\frac{\mathcal{L}_{\kappa}\left(x\right)}{\kappa}.
\label{eq:qhuberloss}
\end{equation}

Finally, the objective of the critic using quantile Huber loss function can be approximated by sampling $K$ and $K'$ independent quantile fractions:
\begin{equation}
\mathcal{L}\left(\phi\right) = \sum^{K}_{k=0}\sum^{K'}_{k'=0}\left(\tau_{k}-\tau_{k'}\right)\rho^{\kappa}_{\hat{\tau_i}}\left(\delta_{kk'}\right).
\label{eq:criticloss}
\end{equation}

\subsubsection{Policy Learning under Distorted Confidence}\label{subsubsec:learnpolicy}
Following the traditional actor-critic framework, the actor's policy is optimized based on the evaluation from the critic.
Although the distributional critic accounts for the uncertainty in the system dynamic, it only approximates an overall impact after the ego agent's decision. Thus the critic can not distinguish the contribution of an ego agent from the contributions of other agents to the current state of the system. To deal with this limitation, we impose distortion weights over the quantiles to represent the confidence in the ego agent's contribution to the overall impact. Intuitively, higher weights on larger quantiles indicate strong confidence in the contribution of ego agent to the overall impact.

Assuming actor is parameterized as $\mu\left(s;\theta\right)$, it can be trained by maximizing the following objective:
 \begin{equation}
\mathcal{J}\left(\theta\right) =  \mathrm{E}\left[\hat{Q}\left(s_t,\mu\left(s_t;\theta\right)\right)\right],
\label{eq:actorloss}
\end{equation}
where the distorted state-action value expectation can be formulated as $\hat{Q}\left(s_t,\mu\left(s_t;\theta\right)\right)\approx \sum^{N-1}_{i=0}\left(\tau_{i+1}-\tau_{i}\right)w\left(\hat{\tau}_i\right)Z_{\hat{\tau}_i}\left(s_t,a_t;\phi\right)$. In Eq.~\eqref{eq:actorloss}, the distortion weights $w\left(\hat{\tau}_i\right),i=0,2,..,N-1$ is predefined. In the next subsection we propose a meta-learning scheme on distortion weights to allow a more flexible and robust learning framework.

 \subsubsection{Robust Multi-agent Control with Meta-Gradient Learning} \label{subsubsec:meta}

\begin{figure}[!t]
\centering
   \includegraphics[scale=0.25]{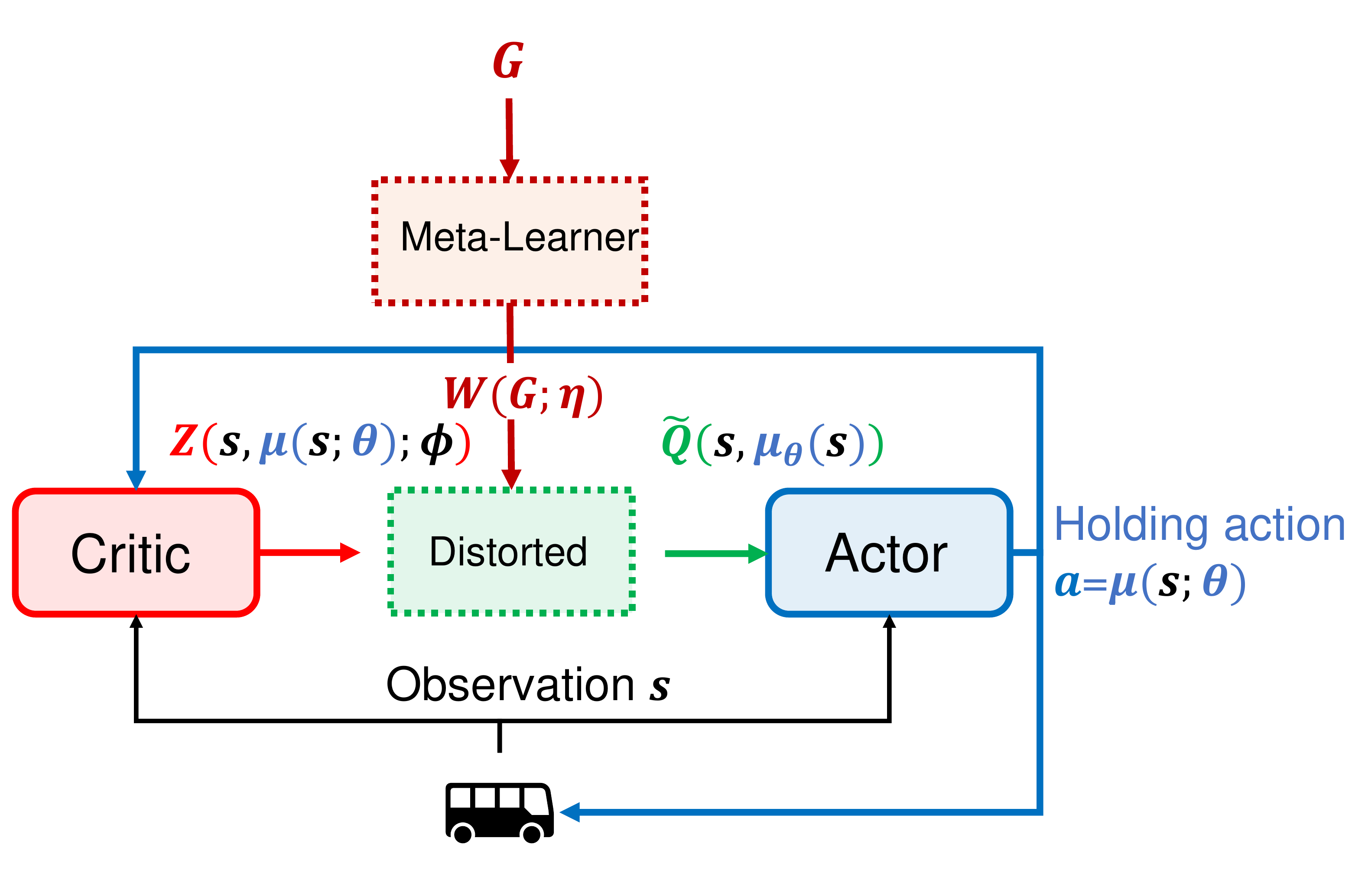}
 \caption{Meta-learning on distortion weights for robust policy learning}
  \label{fig:meta}
\end{figure}

 In this study, meta-learning is introduced to learn adaptive distortion weights to distort state-action value distribution from the critic. To obtain adaptive distortion weights, we utilize the same structure and information similar to the event critic in our previous work \cite{wang2021reducing}. Specifically, the distortion weights are dynamically determined based on the asynchronous events between two consecutive actions of an ego agent (i.e., the holding decisions of other buses). Specifically, the distortion weights are modeled as the output of $W\left(G_t;\eta\right)$, where $W$ represents a graph attention neural network and a set of asynchronous events $G_t$ at decision step $t$ as input (e.g., in \citet{wang2021reducing}, $G_t$ is formulated as a graph to highlight the spatial-temporal relation among each control decision). This scheme is expected to improve the policy learning from two aspects. First, \citet{wang2021reducing} proposed event graph $G_t$ to incorporate asynchronous control events as global information for accurate credit assignment. In this study, the meta-learning module is introduced to learn adaptive distortion weights taking into account the asynchronous events. Therefore we can also incorporate global information to improve policy learning.
 Second, as mentioned in \ref{subsubsec:learnpolicy}, while distributional critic evaluates the impact after agent's action under the uncertainty of the transit system and the behavior of other agents, the adaptive distortion weights can further express how confident the evaluation is in the contribution of the ego agent's action. Therefore by optimizing the meta-learning module, we can distort state-action value distribution towards better policy learning. The overall framework is summarized in Fig.~\ref{fig:meta}. Within this framework we can modify the objective of the actor as:
 \begin{equation}
\mathcal{J}\left(\theta\right) = {\mathrm{E}}\left[\hat{Q}\left(s_t,\mu\left(s_t;\theta\right)\right)\right],
\label{eq:meta_actor}
\end{equation}
where the distorted state-action value expectation is:
 \begin{equation}
  \hat{Q}\left(s_t,\mu\left(s_t;\theta\right)\right)\approx \sum^{N-1}_{i=0}\left(\tau_{i+1}-\tau_{i}\right)W\left(G;\eta\right)_{i}Z_{\hat{\tau}_i}\left(s_t,a_t;\phi\right) .
 \label{eq:meta_Q}
\end{equation}

We follow the meta-policy gradient \cite{metazhong2018} to train $W$ to produce proper distortion weights in $J$ at each actor training step. The meta-objective $J'$ for $W$ can be formulated as:
 \begin{equation}
\mathcal{J}' = {\mathrm{E}}\left[\hat{Q}\left(s_t,\mu\left(s_t;\theta'\right)\right)\right],
\label{eq:meta_obj}
\end{equation}
where $\hat{Q}\left(s_t,\mu\left(s_t;\theta \right)\right)\approx \sum^{N-1}_{i=0}\left(\tau_{i+1}-\tau_{i}\right)Z_{\hat{\tau}_i}\left(s_t,a_t;\phi\right)$. Following \citet{metazhong2018}, in the meta-objective $J'$ we fix the distortion weights to 1. Such setting essentially cross-validates the updated parameters $\theta' = \theta+\alpha \frac{\partial J\left(\theta\right)}{\partial \theta}$, where $\alpha$ is learning rate. It presents an intuitive motivation: the actor just updated by meta-weights should obtain improvement in general distorted state-action value expectation. By using meta distortion weights in $J$, we can achieve more adaptive and more robust policy evaluation to train the actor. Note that the proposed framework achieves continuous control based on IQN and meta-learning, we name it as IQNC-M. To learn robust holding control policy, the details of the proposed training procedure are shown in Algorithm~\ref{alg:A1}.

\begin{algorithm}[!t]
\caption{IQNC-M for Transit Control}
 \label{alg:A1}
\begin{algorithmic}
\STATE Set time horizon $T$ based on real-world operation data.
\STATE Set memory buffers $\left\{\mathcal{B}_i=\emptyset\right\}_{i \in {N}}$, minibatch size as $c$, size to begin training as $B$, number of experience as $C=0$.
\STATE Initialize parameters $\phi$, $\theta$, $\eta$ for actor, critic and meta learner.
\STATE Set learning rate $\alpha_{a}$, $\alpha_{c}$ ,$\alpha_m$ for actor, critic and meta learner, $K$ for the number of quantiles.

\FOR {each episode}
    \STATE {Set agents' last activated time $\left\{t^{-}_i=0\right\}_{i \in {N}}$. }
    \STATE {Set empty memory stacks $\left\{\text{Ms}_i,\text{Ma}_i,\text{Mr}_i\right\}_{i \in {N}}$}
    \FOR {$t =0$ to $T$}
        \FOR {Buses $b_i \in {N} $}
            \IF {Bus $b_i$ arrives at bus stop}
                 \STATE Observe $s_{i,t}$ and compute $a_{i,t}=\mu_{\theta}\left(s_{i,t}\right)$
                 \STATE $\text{Ms}_i \leftarrow \text{Ms}_i \cup  \left\{s_{i,t}\right\}$
                 \STATE $\text{Ma}_i \leftarrow \text{Ma}_i \cup \ \left\{a_{i,t}\right\}$
                 \IF {$t^-_i>0$}
                    \STATE $\text{Mr}_i  \leftarrow \text{Mr}_i  \cup \left\{r_{i,t}\right\}$
                 \ENDIF
                 \IF {$t^-_i>1$}
                     \STATE $g_{i,t^-_i}=\left\{\left(s_{j,t'},a_{j,t'}\right) ,\forall (j,t')\in \mathcal{V}_{i,t^-_i},j\neq i \right\}$

                   \STATE $\mathcal{B}_i \leftarrow \mathcal{B}_i \cup \ \left(a_{i,t^-_i},s_{i,t^-_i},r_{i,t_i},g_{i,t^-_i} \right)$

                   C = C+1
              \ENDIF
              \STATE $t^-_i \leftarrow t$
       \ENDIF
   \ENDFOR
     \STATE Proceed simulation to next step
   \ENDFOR
\IF{$  C  >B$}
   \STATE Sample buffer $\mathcal{B}_i$
    \STATE Sample $c$ experience from $\mathcal{B}_i$

    \STATE Set $K$ quantiles $\tau_{k}= \frac{k}{N} , k=0,...,K$ and $K'$ target quantiles $\tau'_{k'}= \frac{k'}{K'}, k'=0,...,K'$
    \STATE Generate samples $Z_{\tau_k},k=0,...,K$,$Z'_{\tau_k'},k'=0,...,K'$.
    \STATE Sort the samples $Z_{\tau_k}$, $Z_{\tau_k'}$ in ascending order
    \STATE Compute $\delta_{kk'}$ using Eq.~\eqref{eq:td}
     \STATE Update $\phi$ using gradient step $\alpha_{c} \bigtriangledown \mathcal{L}\left(\phi\right)$
    \STATE Update $\theta$ using gradient step $\alpha_{a}\bigtriangledown \mathcal{J}\left(\theta \right)$
    \STATE Update $\eta$ using gradient step $\alpha_{m}\bigtriangledown \mathcal{J'}\left(\eta \right)$
\ENDIF

\ENDFOR
\end{algorithmic}
\end{algorithm}

\section{Experiments} \label{sec:experiments}
\subsection{Experiments Setup}
\subsubsection{Experimental data}
 We evaluate the proposed method based on real-world data. Following the experiment settings and scenarios in \citet{wang2021reducing}, we conduct our experiments on four selected bus routes in an Asian city, with actual passenger demand derived from smart card data. The four routes are all trunk services covering more than 15 km with over 40 bus stops along the route. Table~\ref{lines} lists the basic statistics of the routes, including the number of services per day, the number of bus stops on the route, route length, the mean and standard deviation (std) of headway at the departure terminal.
\begin{table}[!ht]
	\centering
    \caption{Basic information of bus lines.}
    \footnotesize
\begin{tabular}{ccccccc}
\toprule
		 & services  & stops & length (km) & mean (sec) & std (sec)  \\
		\midrule
		R1 & 59  &46 & 17.4 &874 & 302  \\
		R2 & 72 &58 & 23.7 &745 & 307  \\
		R3 & 57  &61  & 23.2 &931 &354 \\
		R4   & 55  &46 & 22.5 &955 &351 \\
		\bottomrule
	\end{tabular}
	\label{lines}
\end{table}

\begin{figure}[!htbp]
\centering
 \includegraphics[scale=0.33]{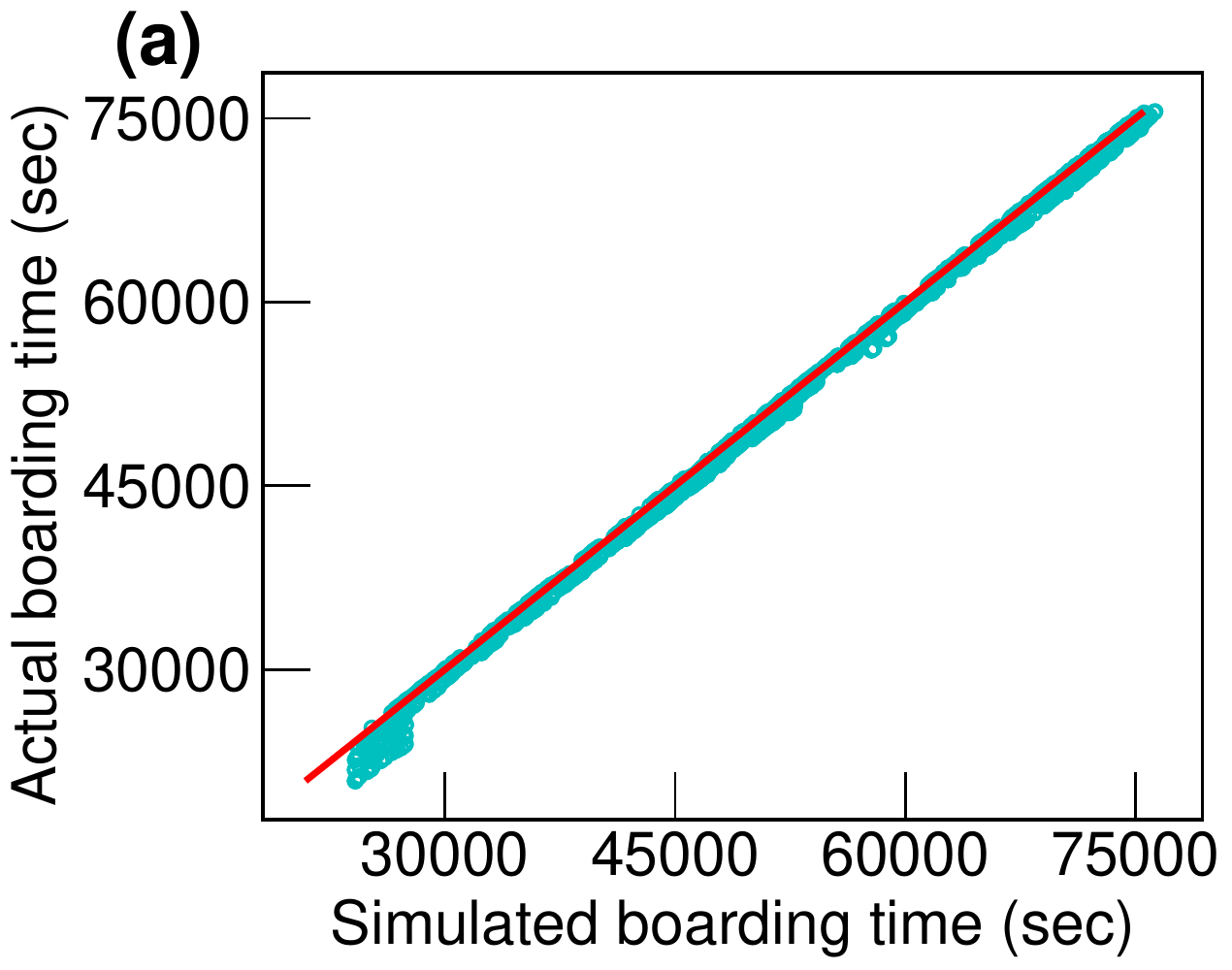} \quad
\includegraphics[scale=0.33]{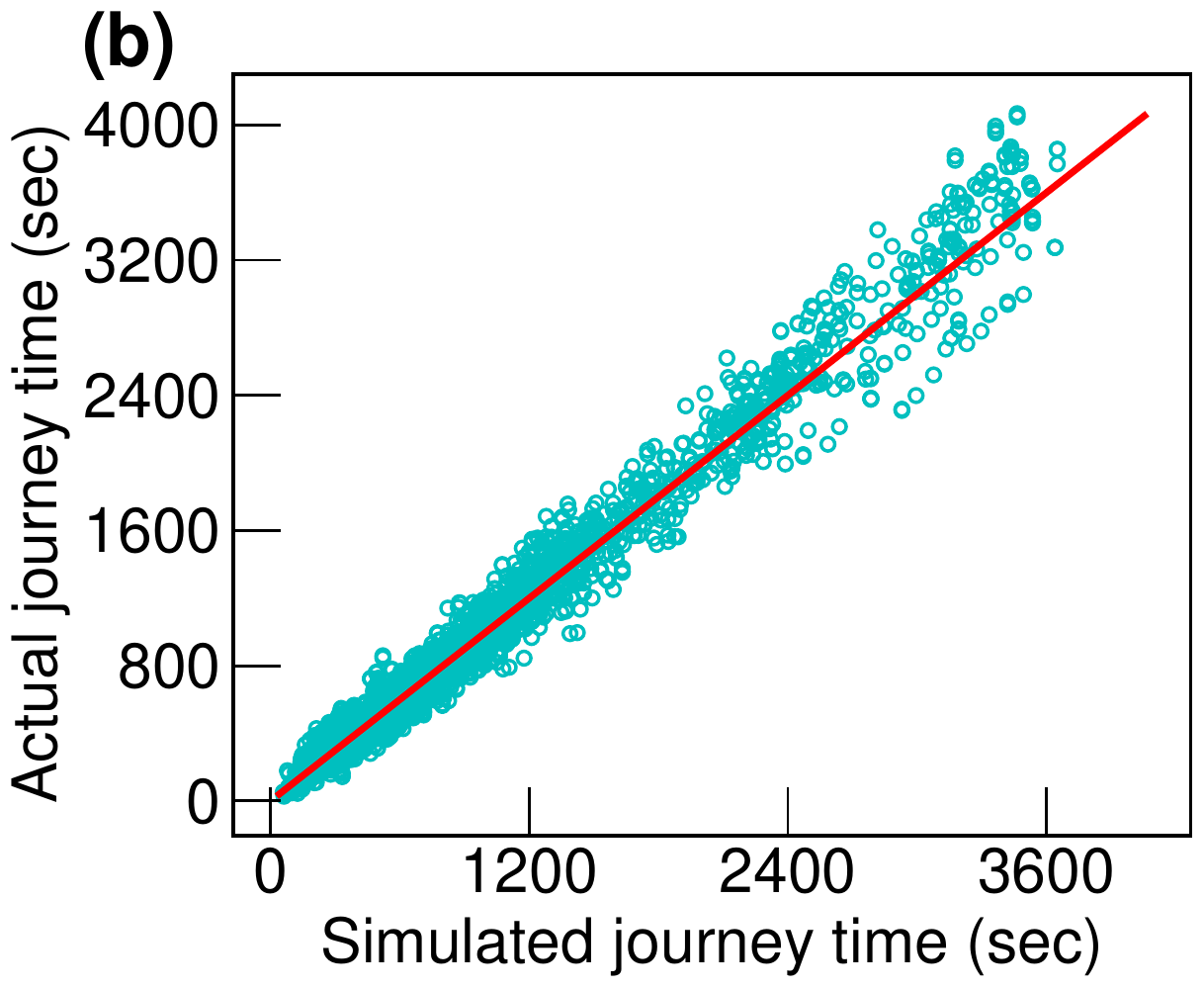}
\caption{Simulated values v.s. true values from smart card data on R1: (a) boarding time (i.e., smart card tapping-in time), and (b) journey time (i.e., duration between tapping-in and tapping-out). }
   \label{sim}
 \end{figure}

We develop and calibrate the bus simulator to reproduce the patterns of real-world operation. In this simulation, the alighting and boarding times per passenger are set to $t_a=1.8\ \text{sec/pax}$ and $t_b=3.0 \ \text{sec/pax}$, respectively. To introduce uncertainties of road conditions, buses are given a random speed $v\times\mathcal{U}\left(0.6,1.2\right)$ km/h when traveling between every two consecutive stops, where $v$ is set to 30 km/h and $\mathcal{U}$ denotes a continuous uniform distribution. The capacity of the bus is set to 120 $\text{pax}$.
Fig.~\ref{sim} (a) and (b) show the simulated boarding time and actual boarding time from smart card (tap-in for boarding and tap-out for alighting) data and the simulated journey time and actual journey time, respectively, for service R1. The Pearson correlations for these two plots are 0.999 and 0.983, respectively, suggesting that our simulator is highly consistent with the real-world operation.

\subsubsection{Baseline models}
The proposed \textbf{IQNC-M} framework is compared with the following baseline models, including both traditional headway-based control models and state-of-the-art RL methods. We also consider variants of the proposed model for validation:
\begin{enumerate}[(i)]
    \item {\textbf{No control (NC)}:} NC is considered as a naive baseline where no holding control is implemented;
    \item {\textbf{Forward headway-based holding control (FH)}: } \cite{daganzo2009headway}: the holding time is $d=\max \left\{ 0, \overline{d}+g\left(H_0-h^-\right) \right\}$, where $H_0$ is the original headway for dispatch, $h^-$ is forward headway of the arriving bus, $\overline{d}$ is the average delay at equilibrium, and $g>0$ is a control parameter. We use the same parameters as in \citet{daganzo2009headway};
    \item {\textbf{Independent Actor-Critic (IAC)}:} \cite{lillicrap2015continuous}: we implement DDPG in the IAC setting to examine the performance where the agent completely overlooks the impact from other agents;
    \item {\textbf{Credit Assignment Framework for Asynchronous Control (CAAC)}:} \cite{wang2021reducing}: we implement CAAC as a state-of-the-art MARL method for transit control;
    \item {\textbf{IQNC-N}:} we implement a continuous version of implicit quantile network (IQN) \citep{dabney2018implicit}, which is a baseline to examine naive distributional RL in a multi-agent setting where there are no distortion weights to indicate the confidence of the policy evaluation;
    \item {\textbf{IQNC-UCF}:} we implement IQN with distortion risk measure as Eq.~\eqref{eq:distort}. This setting indicates that buses are unconfident in their contribution to the transit system along the control horizon. We set $\beta$ in this method is 0.8;
    \item {\textbf{IQNC-CF}:} Similar to IQNC-UCF, we set $\beta=-0.8$ for risk-seeking policy, which can be interpreted as buses are confident in its contribution to the system along the control horizon.
\end{enumerate}

\subsubsection{Evaluation metrics} Following \citep{wang2021reducing}, we use the following indicators to evaluate model performance:
\begin{itemize}
    \item \textbf{Average holding time (AHT)}, which characterizes the degree of intervention;
    \item \textbf{Average waiting time (AWT)}, which evaluates the severity of bus bunching;
    \item \textbf{Average journey time (AJT)}, which quantifies the average journey duration from boarding to alighting for all trips;
    \item \textbf{Average travel time (ATT)}, which quantifies the average travel time for each bus, between the departure terminus to the final terminus;
    \item \textbf{Average occupancy dispersion (AOD)}, which evaluates how balanced the occupancy is. Note the dispersion is computed as a variance-to-mean ratio for the occupancy measures. We expect to see a large AOD value when bus bunching happens, as the loading will be highly imbalanced.
\end{itemize}

\subsubsection{Experimental scenarios}
We conduct robustness analysis by artificially manipulating and disturbing the basic transit demand or traffic state. Specifically, the disturbed transit demand $d_{i,j}(\text{pax/hour})$ with the origin $i$ and the destination $j$ can be modeled as:
\begin{equation}
    d_{i,j} = \hat{d}_{i,j}*p_{d},\  p_{d}\sim \mathcal{N}_{d}(1,\sigma_{d}^2),
\end{equation}
where $\hat{d}_{i,j}$ is the basic demand estimated from smart card data. The scaling factor for demand $p_{d}$ is sampled from a normal distribution $\mathcal{N}_{d}(1,\sigma_{d}^2)$. The standard deviation $\sigma_{d}$ controls the level of uncertainty on the simulated transit demand. Similarly, we can define the disturbed cruising speed between stops $i$ and $j$ as:
\begin{equation}
    s_{i,j} = \hat{s}_{i,j}*p_{s}, \ p_{s}\sim \mathcal{N}_{s}\left(1,\sigma_{s}^2\right),
\end{equation}
where $\hat{s}_{i,j}\left(\text{km/s}\right)$ is the mean cruising speed estimated from smart card data. The scaling factor for speed $p_{s}$ is sampled from a normal distribution $\mathcal{N}_{s}\left(1,\sigma_{s}^2\right)$. The standard deviation $\sigma_{s}$ models the level of uncertainty on the disturbed cruising speed.

To enhance the generalization ability, at each training episode, we impose additional uncertainty by sampling $\sigma_{d}$, $\sigma_{s}$ from a uniform distribution. In this way, we generate different strengths of uncertainties. We conduct experiments with  $\sigma_{d} \sim \mathcal{U}\left[0,3\right]$, $\sigma_{s} \sim \mathcal{U}\left[0,0.3\right]$. 

\subsection{Result Analysis}

 \subsubsection{Evaluation on traffic state perturbations}
We first fix transit demand uncertainty (i.e., $\sigma_{d}=1$ ) and analyze the performance of control strategies by disturbing the normal traffic state with different levels of perturbations (i.e., $\sigma_{s}=\{ 0.1,0.2,0.3\}$). We summarize the results of different models in Table~\ref{table:s4}. As can be seen, in general, stronger traffic state perturbations will lead to less reliable and less efficient transit services. Overall, the proposed IQNC-M achieves the best regularity on the system with a moderate increase in travel time, confirming its robustness under various degrees of traffic state uncertainty. Specifically, both the proposed model and CAAC can achieve stable and efficient control performance, but other RL baseline models essentially fail. This is mainly due to the fact that those RL models without utilizing global information  (i.e., control events of other buses) suffer from unstable training in the dynamic transit system, especially when additional uncertainty is introduced. This result further confirms the importance of uncertainty in training RL models. Notably, we find that the baseline models based on distributional RL methods (i.e., IQNC-N, IQNC-UCF, IQNC-CF) are also less effective than IQNC-M. We believe this is mainly due to the challenges of training distributional RL models in the non-stationary multi-agent setting. By introducing the meta-learning module to incorporate global information, the proposed IQNC-M becomes efficient and effective in addressing this issue. We also find that IQNC-UCF implements the least holding control interventions (with minimum AHD). As IQNC-UCF distorts the action-value distribution to place less confidence on the contribution of an ego bus, it turns out that the learned policy will not encourage a long holding period. Moreover, from the result on R2-R4, we evaluate the RL models on the route in which no training is conducted. Thus the transferability of the RL models can be validated. Though both CACC and the proposed IQNC-M have utilized global information, IQNC-M outperforms CACC in different bus routes operation. We suggest that it is because the meta-learning module produces adaptive distortion weights, which directly results in a better form of state-action value distribution for policy optimization. Although CACC conducts credit assignment to distinguish the contributions of an ego bus from others, this scheme may introduce additional approximation error and hinder policy learning.

 \begin{table*}[!htbp]
\begin{center}
\footnotesize
\begin{tabular}[]{ c c c c c}
\toprule
{Method} &\multicolumn{4}{c}{Execution performance with respect to $\sigma_{s}=\{ 0.1,0.2,0.3\}$ on trained route (R1)} \\
\cmidrule{2-5}
~ & AHD (sec) &$ \Delta$ AWT (sec) & $ \Delta$ ATT (sec) &$ \Delta$ AOD \\
\midrule

NC & - / - / -
& 527 / 548 / 578
& 1019 / 1058 / 1147
& 10.5 / 11.3 / 11.7\\

FH & 57 / 58 / 61
& 1 / -15 / -25
& +575 / +576 / +606
& 0.3 / -0.2 / 0.1\\

CAAC & 33 / 32 / 32
& -31 / -48 / -56
& +284 / +271 / +255
& -2.6 / -3.4 / -3.3\\

IAC & 32 / 31 / 31
& -27 / -43 / -51
& +266 / +253 / +236
& -2.3 / -2.9 / -2.7\\

IQNC-N & 23 / 23 / 22
& -27 / -43 / -38
& +159 / +150 / +150
& -2.2 / -2.9 / -2.3\\

IQNC-UCF & \textbf{22} / \textbf{21} / \textbf{21}
& -26 / -44 / -34
& \textbf{+157} / \textbf{+149} / \textbf{+145}
& -2.1 / -2.9 / -1.7\\

IQNC-CF & 22 / 23 / 23
& -33 / -48 / -53
& +155 / +146 / +139
& -2.4 / -3.0 / -2.7\\

IQNC-M & 37 / 37 / 39
& \textbf{-46} / \textbf{-65} / \textbf{-75}
& +319 / +310 / +320
& \textbf{-3.8} / \textbf{-4.7} / \textbf{-4.7}\\
\midrule
 &\multicolumn{4}{c}{Execution performance with respect to $\sigma_{s}=\{0.1,0.2,0.3\}$ on untrained route (R2)} \\
\cmidrule{2-5}
~ & AHD (sec) &$ \Delta$ AWT (sec) & $ \Delta$ ATT (sec) &$ \Delta$ AOD \\
\midrule
NC & - / - / -
& 553 / 562 / 581
& 1079 / 1113 / 1196
& 17.5 / 17.8 / 18.4\\

FH & 62 / 63 / 65
& -59 / -67 / -67
& +523 / +523 / +541
& -3.2 / -3.5 / -3.8\\

CAAC & 32 / 31 / 31
& -105 / -112 / -103
& +186 / +180 / +176
& -7.0 / -7.3 / -6.5\\

IAC & 31 / 31 / 30
& -94 / -100 / -83
& +180 / +173 / +173
& -6.0 / -6.3 / -5.3\\

IQNC-N & 24 / 23 / 23
& -77 / -80 / -57
& +115 / +110 / +123
& -5.0 / -5.2 / -4.0\\

IQNC-UCF & \textbf{22} / \textbf{22} / \textbf{21}
& -77 / -83 / -66
& \textbf{+104} / \textbf{+100} / \textbf{+110}
& -5.0 / -5.2 / -4.2\\

IQNC-CF & 23 / 23 / 23
& -87 / -91 / -77
& +105 / +101 / +106
& -5.7 / -5.8 / -5.0\\

IQNC-M & 40 / 40 / 40
& \textbf{-127} / \textbf{-130} / \textbf{-129}
& +271 / +267 / +272
& \textbf{-8.5} / \textbf{-8.7} / \textbf{-8.5}\\
\midrule
 &\multicolumn{4}{c}{Execution performance with respect to $\sigma_{s}=\{ 0.1,0.2,0.3\}$ on untrained route (R3)} \\
\cmidrule{2-5}
~ & AHD (sec) &$ \Delta$ AWT (sec) & $ \Delta$ ATT (sec) &$ \Delta$ AOD \\
\midrule

NC & - / - / -
& 571 / 588 / 632
& 1051 / 1085 / 1169
& 9.9 / 10.3 / 11.6\\

FH & 61 / 62 / 64
& +22 / +14 / -9
& +753 / +756 / +771
& +0.3 / +0.1 / -0.8\\

CAAC & 34 / 34 / 33
& -46 / -57 / -69
& +314 / +305 / +294
& -2.1 / -2.3 / -2.8\\

IAC & 33 / 33 / 32
& -38 / -50 / -63
& +295 / +287 / +279
& -1.7 / -2.0 / -2.5\\

IQNC-N & 24 / 24 / 23
& -37 / -47 / -48
& +191 / +182 / +180
& -1.6 / -1.7 / -1.9\\

IQNC-UCF & \textbf{22} / \textbf{23} / \textbf{22}
& -34 / -48 / -53
& \textbf{+186} / \textbf{+179} / \textbf{+180}
& -1.5 / -1.9 / -2.1\\

IQNC-CF & 23 / 24 / 24
& -41 / -52 / -60
& +195 / +189 / +179
& -1.8 / -1.9 / -2.3\\

IQNC-M & 40 / 39 / 40
& \textbf{-55} / \textbf{-68} / \textbf{-94}
& +411 / +402 / +397
& \textbf{-2.6} / \textbf{-3.0} / \textbf{-3.8}\\
\midrule
 &\multicolumn{4}{c}{Execution performance with respect to $\sigma_{s}=\{ 0.1,0.2,0.3\}$ on untrained route (R4)} \\
\cmidrule{2-5}
~ & AHD (sec) &$ \Delta$ AWT (sec) & $ \Delta$ ATT (sec) &$ \Delta$ AOD \\
\midrule

NC & - / - / -
& 580 / 600 / 673
& 788 / 815 / 884
& 6.7 / 7.1 / 8.5\\

FH & 60 / 61 / 64
& -21 / -31 / -59
& +379 / +380 / +388
& -0.5 / -0.7 / -1.5\\

CAAC & 34 / 34 / 33
& -68 / -79 / -103
& +150 / +144 / +134
& -2.2 / -2.4 / -2.9\\

IAC & 33 / 33 / 32
& -62 / -74 / -78
& +143 / +133 / +137
& -1.9 / -2.1 / -2.0\\

IQNC-N & 25 / 24 / 24
& -56 / -65 / -86
& +88 / +82 / +77
& -1.7 / -1.7 / -2.0\\

IQNC-UCF &\textbf{22} / \textbf{22} / \textbf{22}
& -59 / -69 / -81
& \textbf{+78} / \textbf{+71} / \textbf{+77}
& -1.8 / -1.9 / -1.9\\

IQNC-CF & 23 / 23 / 23
& -62 / -70 / -72
& +84 / +82 / +83
& -1.9 / -1.9 / -2.1\\

IQNC-M & 39 / 40 / 41
& \textbf{-74} / \textbf{-90} / \textbf{-128}
& +200 / +197 / +194
& \textbf{-2.6} / \textbf{-2.9} / \textbf{-3.5}\\

\bottomrule
\end{tabular}
\caption{Execution performance under different strength of traffic state perturbations. Model performance is evaluated using: 1) average holding time, and changes in---2) average waiting time ($\Delta$AWT), 3) average travel time ($\Delta$ATT), and 4) average occupancy dispersion ($\Delta$AOD)---w.r.t. baseline NC. The best results are highlighted in bold.}\label{table:s4}
\end{center}
\end{table*}

\subsubsection{Evaluation on transit demand perturbations}
In this subsection, we fix traffic state uncertainty (i.e., $\sigma_{s}=0.1$) and analyze how control strategies perform under different transit demand perturbations (i.e., $\sigma_{d}=\{ 1,2,3\}$). We conduct comparative analysis similar to traffic state perturbations. As shown in Table~\ref{table:d4}, it can also be observed that the stronger perturbations in transit demand place a more challenging scenario to control. The proposed model again performs the best in stabilizing the system (i.e., with smaller AWT and AOD) on both trained and untrained bus routes. It should be pointed out that, similar to the previous result, IQNC-UCF determines less holding period and the bus will experience less average travel time. However, such a strategy does not perfectly match the original goal of system stability in this study. Besides, it will be less efficient given more severe traffic situations (i.e., anomalous events). In future research, we can design a more comprehensive reward function to balance the trade-off between dwell time serving passengers and the additional holding period.

 \begin{table*}[!htbp]
\begin{center}
\footnotesize
\begin{tabular}[]{ c c c c c}
\toprule
{Method} &\multicolumn{4}{c}{Execution performance with respect to $\sigma_{d}=\{1,2,3\}$ on trained route (R1)} \\
\cmidrule{2-5}
~ & AHD (sec) &$ \Delta$ AWT (sec) & $ \Delta$ ATT (sec) &$ \Delta$ AOD \\
\midrule
NC & - / - / -
& 527 / 560 / 564
& 1019 / 1071 / 1113
& 10.5 / 13.9 / 15.7\\

FH & 57 / 58 / 58
& +1 / -21 / -17
& +575 / +551 / +566
& +0.3 / -1.5 / -0.3\\

CAAC & 33 / 32 / 32
& -31 / -60 / -59
& +284 / +263 / +230
& -2.6 / -4.0 / -4.5\\

IAC & 32 / 31 / 31
& -27 / -56 / -53
& +266 / +248 / +216
& -2.3 / -3.6 / -3.9\\

IQNC-N & 22 / 23 / 23
& -27 / -56 / -53
& +159 / +144 / +121
& -2.2 / -3.5 / -3.7\\

IQNC-UCF &\textbf{22} / \textbf{22} / \textbf{22}
& -26 / -52 / -48
& \textbf{+157} / \textbf{+144} / \textbf{+118}
& -2.1 / -3.3 / -3.6\\

IQNC-CF & 23 / 23 / 23
& -33 / -61 / -59
& +155 / +145 / +122
& -2.4 / -3.7 / -4.0\\

IQNC-M & 37 / 37 / 37
& \textbf{-46} / \textbf{-76} / \textbf{-80}
& +319 / +297 / +260
& \textbf{-3.8} / \textbf{-5.5} / \textbf{-6.0}\\
\midrule
{Method} &\multicolumn{4}{c}{Execution performance with respect to $\sigma_{d}=\{1,2,3\}$ on untrained route (R2)} \\
\cmidrule{2-5}
~ & AHD (sec) &$ \Delta$ AWT (sec) & $ \Delta$ ATT (sec) &$ \Delta$ AOD \\
\midrule

NC & - / - / -
& 553 / 578 / 656
& 1079 / 1117 / 1151
& 17.5 / 18.9 / 19.3\\

FH & 62 / 62 / 62
& -59 / -55 / -52
& +523 / +494 / +484
& -3.2 / -3.3 / -2.4\\

CAAC & 32 / 31 / 32
& -105 / -102 / -87
& +186 / +169 / +169
& -7.0 / -7.0 / -5.8\\

IAC & 31 / 31 / 31
& -94 / -82 / -69
& +180 / +168 / +170
& -6.0 / -5.8 / -4.6\\

IQNC-N & 24 / 25 / 26
& -77 / -66 / -47
& +115 / +116 / +137
& -5.0 / -4.6 / -3.4\\

IQNC-UCF &\textbf{22} / \textbf{22} / \textbf{22}
& -77 / -66 / -57
& \textbf{+104} / \textbf{+95} / \textbf{+104}
& -5.0 / -4.8 / -3.9\\

IQNC-CF & 23 / 24 / 25
& -87 / -71 / -57
& +105 / +106 / +126
& -5.7 / -5.3 / -4.2\\

IQNC-M & 40 / 40 / 41
& \textbf{-127} / \textbf{-121} / \textbf{-103}
& +271 / +252 / +260
& \textbf{-8.5} / \textbf{-8.3} / \textbf{-6.8}\\
\midrule
{Method} &\multicolumn{4}{c}{Execution performance with respect to $\sigma_{d}=\{1,2,3\}$ on untrained route (R3)} \\
\cmidrule{2-5}
~ & AHD (sec) &$ \Delta$ AWT (sec) & $ \Delta$ ATT (sec) &$ \Delta$ AOD \\
\midrule
NC & - / - / -
& 571 / 653 / 816
& 1051 / 1082 / 1115
& 9.9 / 11.0 / 11.3\\

FH & 61 / 61 / 61
& +22 / +25 / +62
& +753 / +732 / +713
& +0.3 / +0.3 / +0.7\\

CAAC & 34 / 34 / 34
& -46 / -55 / -30
& +314 / +295 / +291
& -2.1 / -2.2 / -1.6\\

IAC & 33 / 33 / 33
& -38 / -45 / -20
& +295 / +282 / +282
& -1.7 / -1.9 / -1.3\\

IQNC-N & 24 / 24 / 26
& -37 / -41 / -29
& +191 / +195 / +216
& -1.6 / -1.8 / -1.2\\

IQNC-UCF & \textbf{23} / \textbf{23} / \textbf{23}
& -34 / -43 / -31
& \textbf{+186} / \textbf{+177} / \textbf{+184}
& -1.5 / -1.8 / -1.1\\

IQNC-CF & 24 / 24 / 25
& -41 / -49 / -33
& +195 / +198 / +219
& -1.8 / -2.0 / -1.3\\

IQNC-M & 40 / 39 / 39
& \textbf{-55} / \textbf{-63} / \textbf{-36}
& +411 / +389 / +375
& \textbf{-2.6} / \textbf{-2.8} / \textbf{-1.9}\\
\midrule
{Method} &\multicolumn{4}{c}{Execution performance with respect to $\sigma_{d}=\{1,2,3\}$ on untrained route (R4)} \\
\cmidrule{2-5}
~ & AHD (sec) &$ \Delta$ AWT (sec) & $ \Delta$ ATT (sec) &$ \Delta$ AOD \\
\midrule
NC & - / - / -
& 580 / 609 / 738
& 788 / 820 / 853
& 6.7 / 7.8 / 9.3\\

FH & 60 / 60 / 61
& -21 / -8 / -20
& +379 / +369 / +351
& -0.5 / -0.6 / -0.9\\

CAAC & 34 / 34 / 33
& -68 / -72 / -92
& +150 / +137 / +128
& -2.2 / -2.2 / -2.4\\

IAC & 33 / 33 / 33
& -62 / -66 / -86
& +143 / +131 / +123
& -1.9 / -1.9 / -2.1\\

IQNC-N & 25 / 25 / 27
& -56 / -58 / -70
& +88 / +83 / +87
& -1.7 / -1.6 / -1.7\\

IQNC-UCF &\textbf{22} / \textbf{22} / \textbf{22}
& -59 / -53 / -77
& \textbf{+78} / \textbf{+74} / \textbf{+70}
& -1.8 / -1.6 / -1.6\\

IQNC-CF & 23 / 24 / 25
& -62 / -67 / -87
& +84 / +79 / +86
& -1.9 / -1.8 / -1.9\\

IQNC-M & 39 / 40 / 40
& \textbf{-74} / \textbf{-79} / \textbf{-105}
& +200 / +187 / +176
& \textbf{-2.6} / \textbf{-2.7} / \textbf{-3.0}\\
\bottomrule
\end{tabular}

\caption{Execution performance under different strength of transit demand perturbations. Model performance is evaluated using: 1) average holding time, and changes in---2) average waiting time ($\Delta$AWT), 3) average travel time ($\Delta$ATT), and 4) average occupancy dispersion ($\Delta$AOD)---w.r.t. baseline NC. The best results are highlighted in bold.}\label{table:d4}
\end{center}
\end{table*}



 \subsubsection{Evaluation on anomaly}
\begin{figure}[!htbp]
\centering
   \includegraphics[scale=0.35]{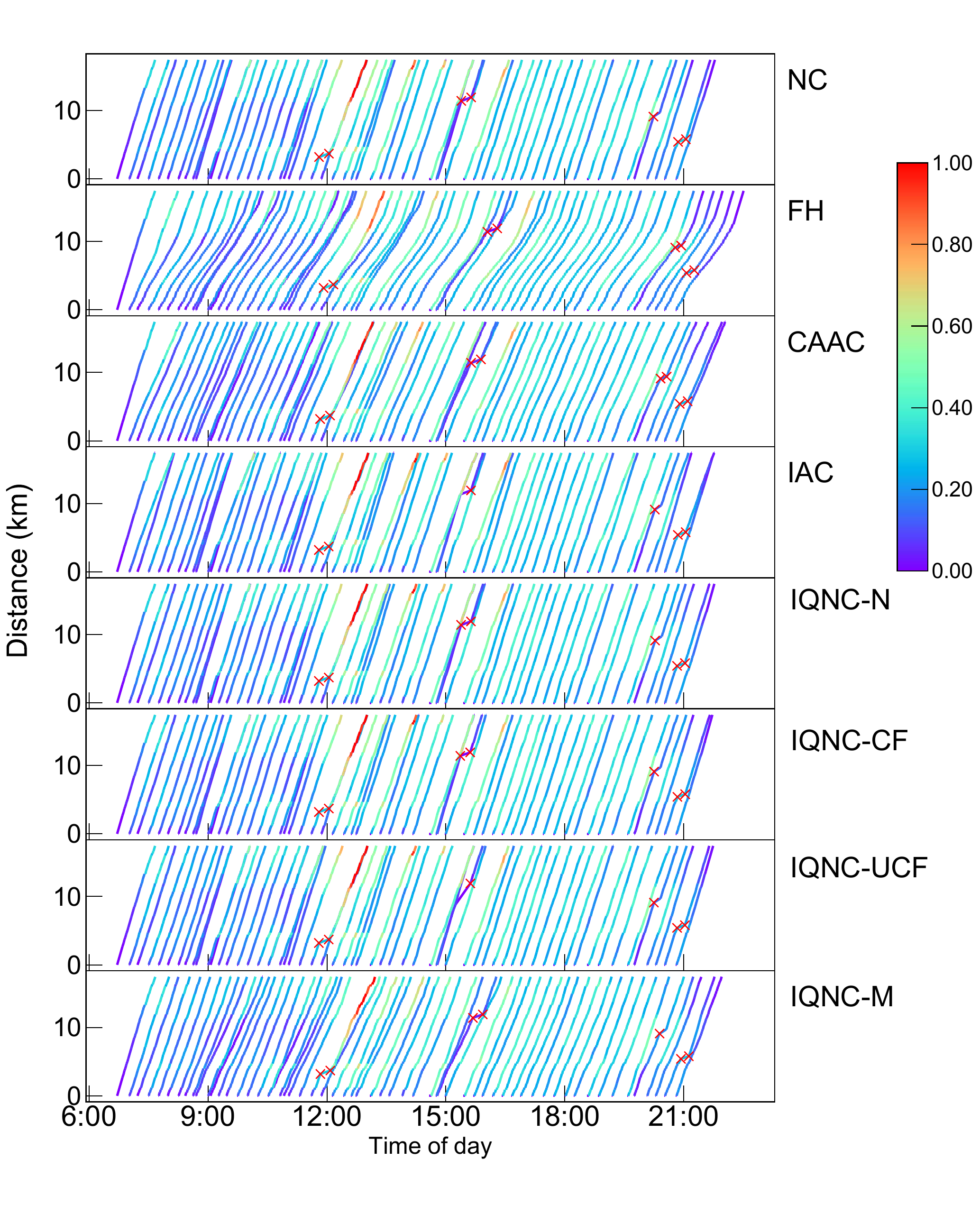}
 \caption{Bus trajectories encountering traffic interruption under different holding strategies.}
  \label{fig:tri}
\end{figure}

\begin{figure}[!htbp]
\centering
   \includegraphics[scale=0.35]{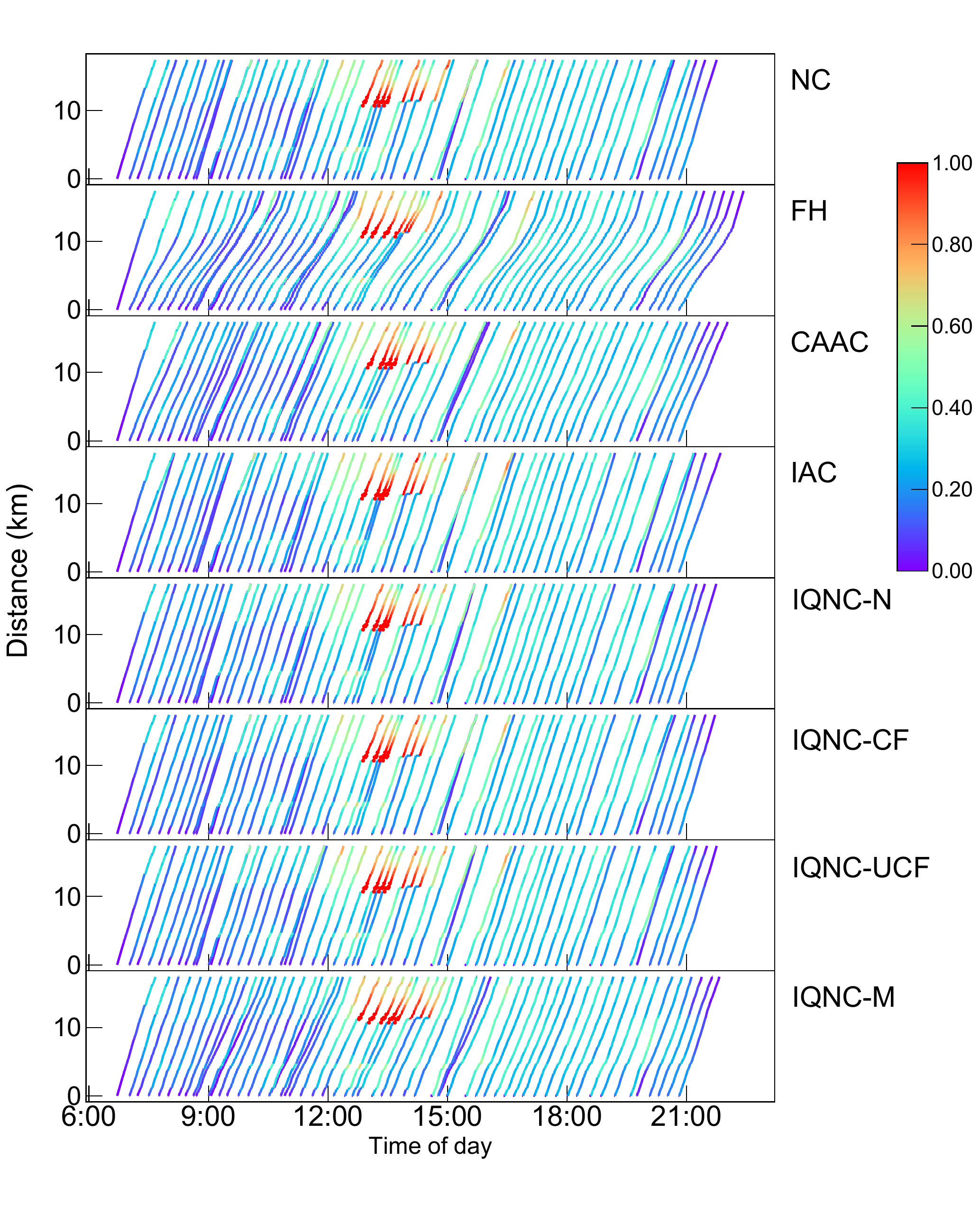}
 \caption{Bus trajectories encountering demand surge under different holding strategies.}
  \label{fig:trd}
\end{figure}

 Apart from the perturbations in the transit system, the anomaly is another tricky issue in daily transit operation resulting from some special events. While most studies overlook such scenarios, we particularly examine the control strategies under two of the most common anomaly scenarios (i.e., demand surges and traffic interruptions). Fig.~\ref{fig:tri} and Fig.~\ref{fig:trd} show the trajectory plot colored by occupancy (i.e., number of onboard/capacity) under different control policies based on the same random seed in traffic interruption and demand surge scenarios, respectively.
 In Fig.~\ref{fig:tri}, we randomly select four buses suffering traffic interruption (i.e., the speed is scaled by 0.1 in this setting), which may be caused by a traffic accident. As can be seen, these abnormal situations provide immediate severe perturbations on the system, which tend to ruin system stability if there is no any intervention (i.e., NC). Notably, the proposed IQNC-M model presents better control interventions to recover and maintain system performance and stability.
 In Fig.~\ref{fig:trd}, we randomly select three stops and impose additional 50 passengers on each bus arrival for a one-hour span, which simulates the demand surge due to some special events in the neighboring area of these stops (i.e., music concert). From the trajectories, we can observe that the proposed IQNC-M achieves a faster recovery of the system regularity after demand surge and maintain efficient control along the following horizon.

\subsubsection{Analysis on meta distortion weights}
\begin{figure}[!t]
\centering
 {\includegraphics[scale=0.4 ]{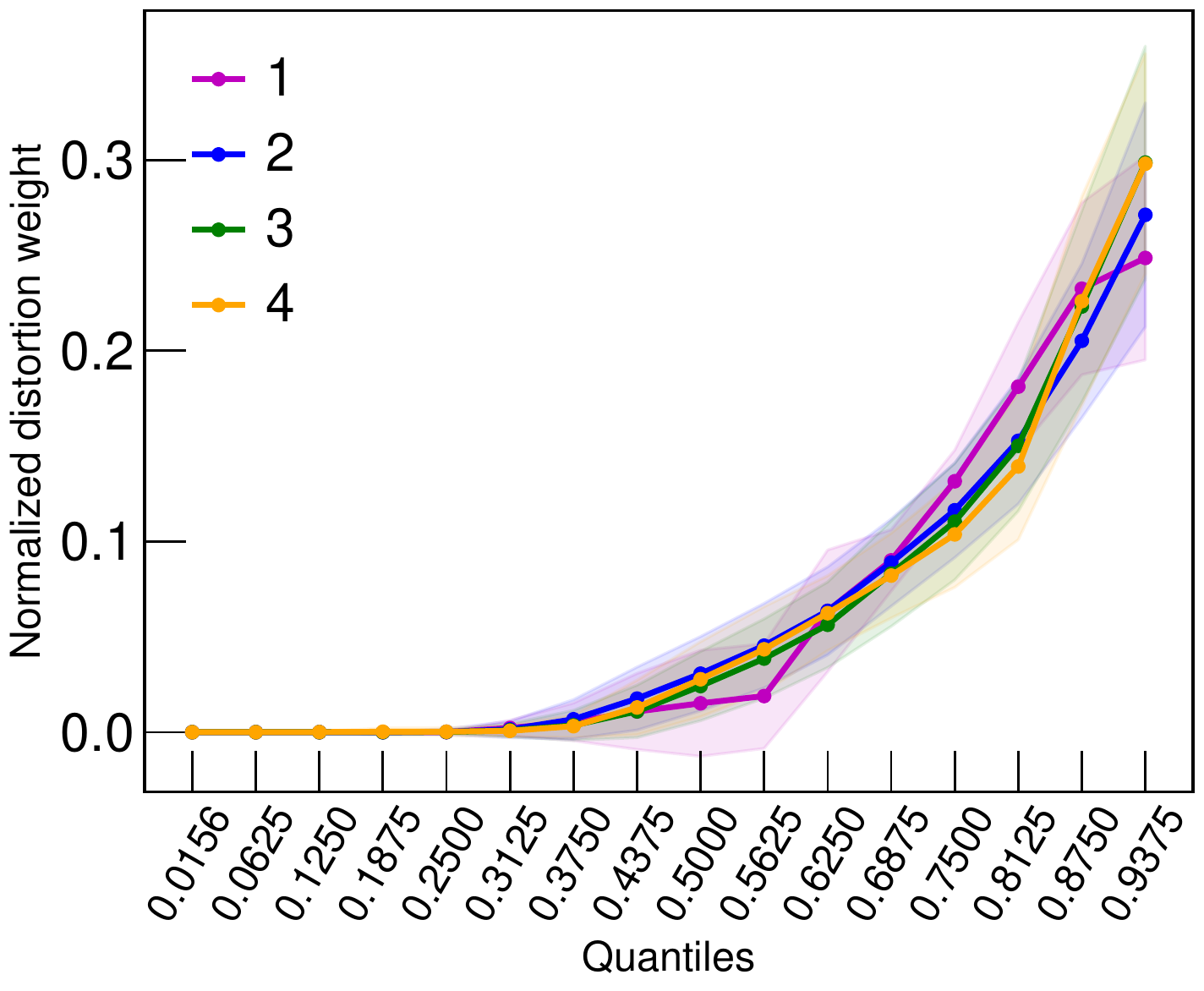}}
\caption{Meta distortion weight at training episode 1}
\label{fig:meta1}
\end{figure}

\begin{figure}[!t]
\centering
 {\includegraphics[scale=0.4 ]{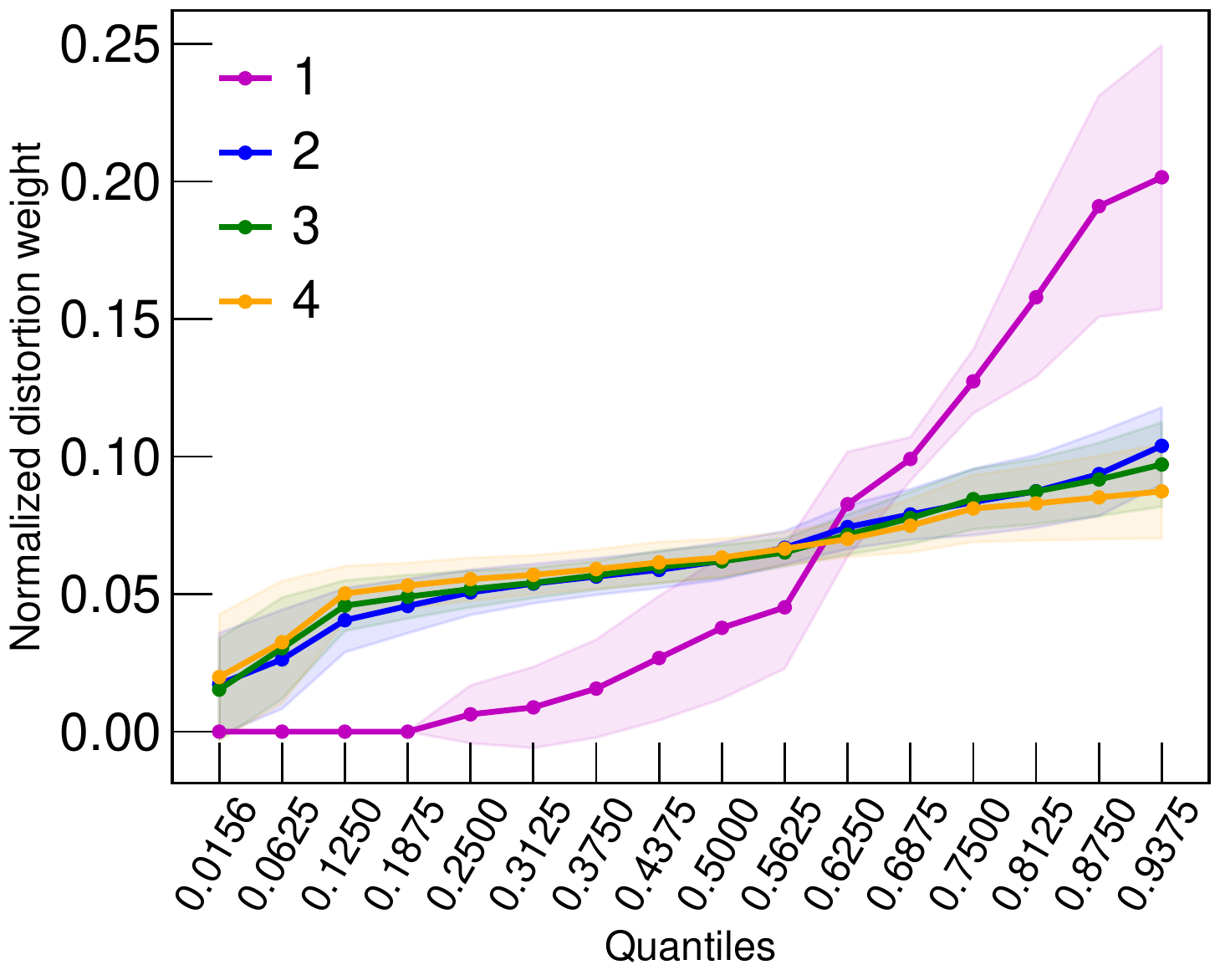}}
\caption{Meta distortion weight at training episode 300}
\label{fig:meta300}
\end{figure}

One of the key contributions in this study is that we design a meta-learning scheme to generate adaptive distortion weights based on global information. In this way, we achieve efficient distributional RL in a multi-agent transit control setting.
While the previous results have demonstrated the efficiency of the proposed meta-learning module, this subsection will further illustrate its interpretability. As shown in Fig.~\ref{fig:meta1} and Fig.~\ref{fig:meta300}, the distortion weights on different quantiles are visualized, which are obtained using multiple random seeds. Specifically, each line indicates a set of distortion weights under a specific number of events (i.e., one indicates only the ego bus implements holding action during its two consecutive stop arrival).
Specifically, at the beginning of the training, the distortion weights corresponding to a different number of events do not present meaningful information since they are nearly the same. In contrast, after 300-episode training, we notice that the distortion weights corresponding to a single event increase sharply for larger quantiles. Meanwhile, there is less increase in distortion weights corresponding to more events. This phenomenon demonstrates that the meta-learning module places higher confidence on the contribution of the ego bus when there are fewer other control events. It makes sense because if there are no other control events between two consecutive holding decisions of the ego bus, its holding should have a more dominant impact on the dynamic of the system. Therefore the distortion weights should give higher credit on its action.

\section{Conclusions and Future Work} \label{sec:conclusion}
In this paper, we propose a distributional MARL framework to achieve robust bus holding control in the daily operation of a bus fleet/route. The proposed IQNC-M framework exploits the distributional RL setting to handle the uncertainties in the transit system. We introduce an interpretable meta-learning module  to achieve efficient distributional RL in a non-stationary multi-agent system. Specifically, we introduce distortion weights to distort the state-action value distribution of the critic, and derive the weights based on the meta-learning module. By doing so, the policy evaluation becomes more robust and efficient for policy training. Finally, we design a training procedure for transit operation for robust control policy learning. Our experiments on real-world bus services demonstrate that the proposed framework can achieve better and more stable control under various perturbations and anomalies.

There are several directions for future research. First, adversarial reinforcement learning could be one of the promising schemes to enhance the policy robustness \citep{pan2019risk}. For example, we could model transit system as an adversary, which is trained to present rational and systematic perturbations/interruptions for the bus agents to develop a more robust control policy. Second, we can extend the framework for multi-line operation, in which different bus routes will share the same road and bus stops. In this case, deriving coordinated control policy among different routes will help achieve robust network-wise operations. Finally, we believe that this study can also contribute to other applications beyond transit operation. The idea of learning adaptive distortion weights for distribution MARL offers a promising solution to achieve efficient learning while sidestepping the credit-assignment challenge for a general multi-agent system.



\ifCLASSOPTIONcaptionsoff
 \newpage
\fi



%

\bibliographystyle{IEEEtranN}
\bibliography{ref}

%

\begin{IEEEbiography}[{\includegraphics[width=1in,height=1.25in,clip,keepaspectratio]{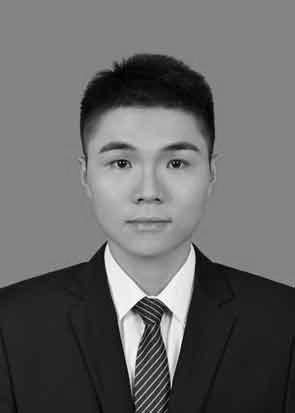}}]
{Jiawei Wang} received the B.S. degree in Traffic Engineering from Sun Yat-Sen University, Guangzhou, China, in 2016, and M.S. degree in Traffic information and control from Sun Yat-Sen University, Guangzhou, China, in 2019. He is now a Ph.D. candidate supervised by Prof. Lijun Sun, with the Department of Civil Engineering at McGill University.
His current research centers on Intelligent transportation systems, traffic control and machine learning.
\end{IEEEbiography}

\begin{IEEEbiography}[{\includegraphics[width=1in,height=1.25in,clip,keepaspectratio]{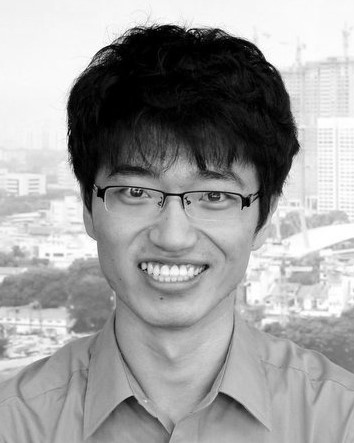}}]
{Lijun Sun} (Member, IEEE) received the B.S. degree in Civil Engineering from Tsinghua University, Beijing, China, in 2011, and Ph.D. degree in Civil Engineering (Transportation) from the National University of Singapore in 2015. He is currently an Assistant Professor with the Department of Civil Engineering at McGill University, Montreal, Quebec, Canada. His research centers on intelligent transportation systems, machine learning, spatiotemporal modelling, travel behavior, and agent-based simulation.
\end{IEEEbiography}




\end{document}